\PassOptionsToPackage{dvipsnames,table}{xcolor}
\documentclass{applemlr}

\usepackage{amsmath}
\usepackage{enumerate}
\usepackage{algorithm}
\usepackage{algpseudocode}
\usepackage{amsfonts}
\usepackage{amsthm}
\usepackage{cleveref}
\usepackage{diagbox}
\usepackage{colortbl}
\usepackage{amssymb}
\usepackage{xspace}
\usepackage{wrapfig}
\usepackage{adjustbox}
\usepackage{tabularx}
\usepackage{booktabs}
\usepackage{mathtools}
\usepackage{tikz}
\usepackage{enumitem}
\usepackage{silence}
\usepackage{dsfont}
\usepackage[table]{xcolor}
\usepackage[dvipsnames]{xcolor}
\usepackage{multirow}
\usepackage{makecell}
\usepackage{xfakebold}

\usepackage{bm}

\DeclareMathAlphabet{\mathsfit}{\encodingdefault}{\sfdefault}{m}{sl}
\SetMathAlphabet{\mathsfit}{bold}{\encodingdefault}{\sfdefault}{bx}{n}

\def\va{{\bm{a}}}
\def\vb{{\bm{b}}}
\def\ve{{\bm{e}}}
\def\vw{{\bm{w}}}
\def\vx{{\bm{x}}}
\def\vy{{\bm{y}}}

\newcommand{\E}{\mathbb{E}}
\DeclareMathOperator*{\argmin}{arg\,min}

\definecolor{textgray}{HTML}{6E6E73}
\usetikzlibrary{positioning, calc}
\usetikzlibrary{decorations.pathmorphing}

\makeatletter
\patchcmd{\wrong@fontshape}{\@gobbletwo}{}{}{}
\makeatother
\numberwithin{equation}{section}
\makeatletter
\AtBeginDocument{
  \urlstyle{sf}
  
}
\makeatother

\definecolor{light}{RGB}{125, 125, 125}
\crefname{tcb@cnt@pbox}{code}{code}
\Crefname{tcb@cnt@pbox}{Code}{Code}
\crefname{assumption}{assumption}{assumption}
\Crefname{assumption}{Assumption}{Assumptions}

\newtcolorbox[auto counter]{pbox}[2][]{
  colback=white,
  title=Code~\thetcbcounter: #2,
  #1,fonttitle=\sffamily,
  fontupper=\sffamily,
  arc=2pt,
  colframe=bgcolor,
  coltitle=fgcolor,
  colbacktitle=bgcolor,
  toptitle=0.25cm,
  bottomtitle=0.125cm
}

\makeatletter
\newcommand\applefootnote[1]{%
  \begingroup
  \renewcommand\thefootnote{}%
  \renewcommand\@makefntext[1]{\noindent##1}%
  \footnote{#1}%
  \addtocounter{footnote}{-1}%
  \endgroup
}
\makeatother

\definecolor{cverbbg}{gray}{0.90}

\usepackage{rotating}
\usepackage{float}

\providecommand{\method}{{HyperTransport}\xspace}
\newcommand{\methodshort}{{HT}\xspace}
\newcommand{\vlmmodel}{InternVL2.5-26B\xspace}

\newcommand{\eg}{\textit{e.g.,}\xspace}
\newcommand{\ie}{\textit{i.e.,}\xspace}

\newcommand{\perceiver}{Perceiver IO\xspace}
\newcommand{\lineas}{\texttt{LinEAS}\xspace}

\newcommand{\conceptfidelity}{\textit{Concept Fidelity}\xspace}
\newcommand{\promptfidelity}{\textit{Input Fidelity}\xspace}
\newcommand{\conceptfidelityshort}{\textit{Concept Fid.}\xspace}
\newcommand{\promptfidelityshort}{\textit{Input Fid.}\xspace}

\newcommand{\actv}{\va}
\newcommand{\crep}{\ve}

\definecolor{trainable}{RGB}{51, 102, 204}

\newcommand{\Model}{\mathsf{G}}
\newcommand{\Enc}{\mathsf{E}}
\newcommand{\Hyperpsi}{{\color{trainable}\psi}}
\newcommand{\Hyper}{{\color{trainable}\mathsf{H}_\Hyperpsi}}
\newcommand{\Interv}{\mathcal{I}}
\newcommand{\src}{\mathrm{src}}

\newcommand{\dmdtwo}{DMD2\xspace}
\newcommand{\nitro}{{Nitro-1-PixArt}\xspace}

\newcommand{\myurl}[1]{{\urlstyle{sf}\tiny\url{#1}}}

\usepackage{pgfplots}
\pgfplotsset{compat=1.18}
\usetikzlibrary{positioning,arrows.meta,shapes,fit,backgrounds,calc}
\usepackage{url}

\definecolor{rebuttalcolor}{RGB}{0, 0, 200}

\newif\ifshowrevisions \showrevisionstrue
\ifshowrevisions\else

\fi

\theoremstyle{plain}

\theoremstyle{definition}

\theoremstyle{remark}

\newcommand{\talldiamond}{\raisebox{-0.1ex}{\scalebox{0.6}[1.1]{$\,\diamond$}}}

\title{HyperTransport: Amortized Conditioning of T2I Generative Models}

\author[1,2,\talldiamond]{Valentino Maiorca}
\author[2]{Eleonora Gualdoni}
\author[2]{Xavier Suau}
\author[2]{Marco Cuturi}
\author[2]{Luca Zappella}
\author[2]{Pau Rodríguez}

\affiliation[1]{ISTA}
\affiliation[2]{Apple}
\affiliation[\talldiamond]{\textit{Work done as intern at Apple}}

\abstract{
As foundation models grow in capability, the ability to efficiently and reliably control their behavior becomes critical. Fine-tuning these models can be costly, and while prompting can be practical for controllability, it remains fragile due to models' high sensitivity to exact prompt wording and structure. This brittleness has driven interest in activation steering techniques that offer more stable and predictable control over model behavior.
However, existing activation steering methods require per-concept optimization, which makes them ill-suited to deployment scenarios where the concept set is large, evolving, or only specified at request time: each new concept incurs at least minutes of optimization on the target model.
We propose \method{}, a hypernetwork framework that amortizes this cost by mapping embeddings from a pretrained encoder (CLIP in our instantiation) directly to intervention parameters, trained end-to-end using an optimal transport loss. 
Once trained, \method{} produces each new intervention in a single hypernetwork forward pass, $3600$--$7000\times$ faster than per-concept fitting. On concepts unseen during training, it matches the strongest per-concept baselines at inducing the target concept. By decoupling concept representation from intervention prediction, \method{} combines three capabilities that no existing approach offers as a set: amortized steering for open-ended concept sets, continuous interpretable strength control, and cross-modal conditioning where reference images can directly steer text-based generation. We validate \method{} on \dmdtwo and \nitro across 167 held-out test concepts via CLIP-based metrics, a VLM-as-a-judge evaluation, and a user study. In pairwise comparisons, both human and VLM judges prefer \method{} over prompting ${\sim}2{\times}$ as often.
}

\metadata[Correspondence]{\sffamily Valentino Maiorca: \url{valentino@maiorca.xyz}; Pau Rodríguez: \url{pau.rodriguez@apple.com}}
\date{\sffamily\today}

\begin{document}

\maketitle

\begin{figure}[htb]
    \centering
    \includegraphics[width=\linewidth]{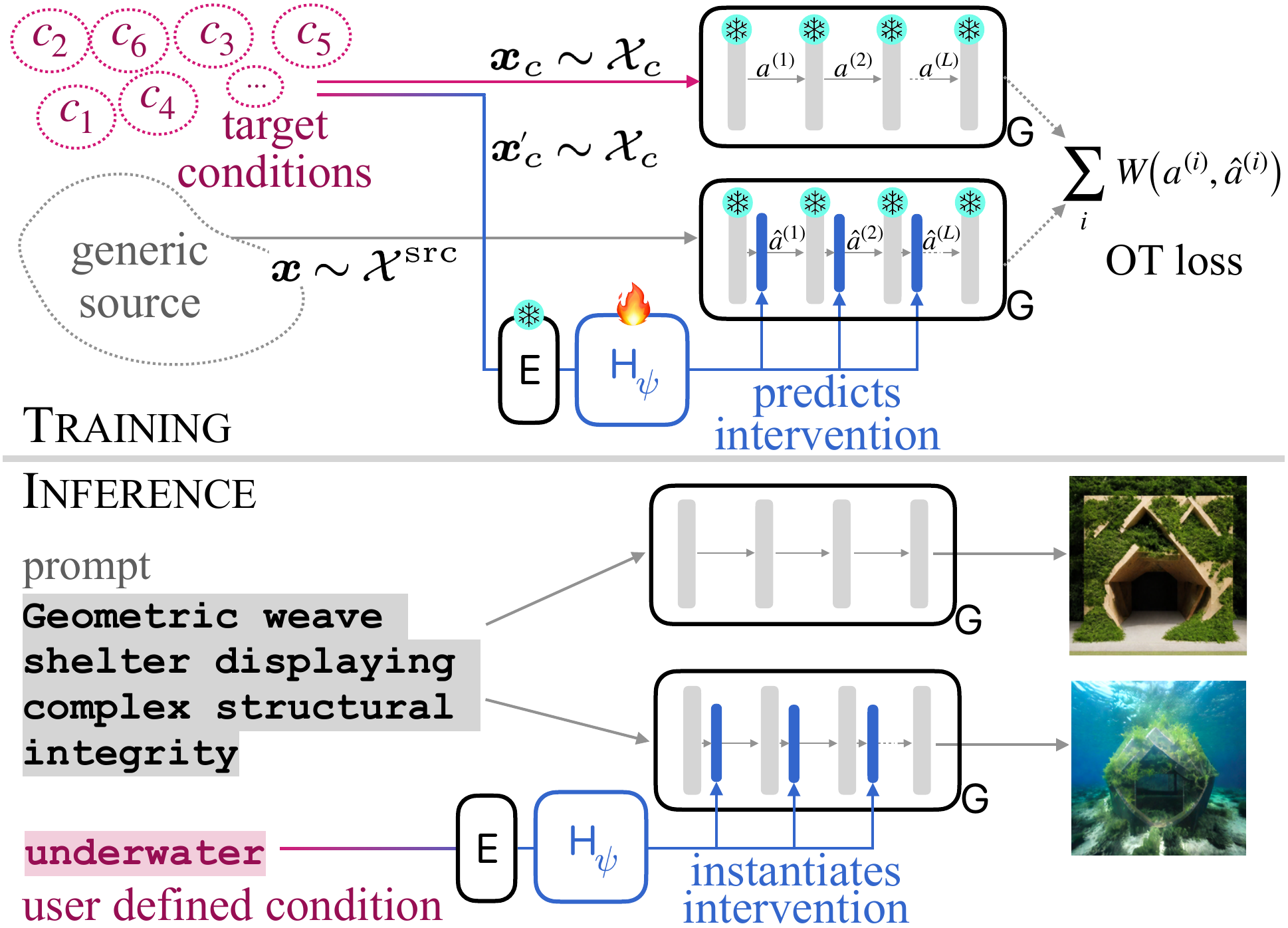}
    \caption{\textbf{Training and inference of \method}. During training (top), we sample $\{\vx_c\}$ sentences corresponding to a concept $c$, along with $\{\vx\}$ unrelated sentences. The hypernetwork $\Hyper$ predicts an intervention on $\Model$ from a target encoding given by an encoder $\Enc$. $\Hyper$ is updated using an Optimal Transport loss $W$ until convergence. During inference (bottom) $\Hyper\circ\Enc$ dynamically predicts high quality interventions for test concepts in negligible time.}
    \label{fig:fig1}
\end{figure}

\section{Introduction}
\label{sec:introduction}

As generative models continue to scale, methods for efficiently controlling them toward specialized behaviors are critical, both for aligning towards human preferences (\eg mitigating toxic language) and for creative applications like personalized image generation.
Traditional approaches rely on fine-tuning entire models~\citep{ouyang2022training} or crafting precise prompts~\citep{brown2020language}, methods that either require substantial computational resources or depend on brittle prompt engineering.

Activation steering takes a different approach: small, targeted modifications of internal activations express a target concept without changing the model's weights. The premise is now well supported~\citep{zou2023representation}: semantic modifications like ``\textit{apply Van Gogh style}'' are identifiable within latent space and reachable through simple, often affine, activation transformations. Recent methods achieve state-of-the-art results across a wide range of concepts, including toxicity mitigation~\citep{lineas}, object removal in 1-step distilled image generators~\citep{linearact}, and truthfulness induction~\citep{li2024inference}. But every new concept still requires \textit{ad-hoc optimization} on the target model, costing minutes of forward (and optionally backward) passes. This rules out interactive interfaces where users specify concepts on demand, content pipelines whose policy set expands over time, and any system that serves unforeseen concepts at query time.

\method (\methodshort) \textit{amortizes} this per-concept optimization through a hypernetwork $1000\times$ smaller than the target image generator. Conditioned on CLIP~\citep{radford2021clipg} representations, it maps directly to intervention parameters and serves any concept well-represented in CLIP space, even ones unseen at training, in a single forward pass. \method thus borrows the activation-space precision of per-concept steering and the open, immediate interface of prompting.

We train the hypernetwork end-to-end through the target generative model, minimizing a 1D-Wasserstein activation alignment loss, the same objective LinEAS~\citep{lineas} uses to achieve state-of-the-art per-concept steering (\Cref{fig:fig1}). Our objective only needs unpaired data, which is far more realistic and cheaper to gather across hundreds of concepts than paired alternatives, and gives \methodshort the bounded, interpretable strength control inherent of optimal-transport steering (\Cref{sec:control}). Training on $610$ concepts takes ${\sim}64$ GPU-hours, $2.7\times$ less than fitting LinEAS to each one individually. On image generation with \dmdtwo~\citep{yin2024dmd2} and \nitro (a one-step distillation of \citet{chen2024pixartsigma}), \method matches the strongest baseline (\ie, \lineas) on 167 globally-applied visual concepts unseen by it at training time but individually fitted by each per-concept baseline, isolating amortization from reachability. 

Decoupling concept representation from intervention prediction unlocks a novel capability closed to both traditional steering and prompting: \textbf{cross-modal conditioning}. The hypernetwork operates on encoder outputs, so any modality the encoder accepts can drive the intervention. With CLIP's multimodal space as input, this is zero-shot at deployment: a reference image can directly condition text-to-image generators with no retraining (\Cref{sec:exp_crossmodal}). 

Our contributions can be summarized as follows:

\begin{itemize}[nosep,leftmargin=*]
\item \textbf{\method Framework.} A hypernetwork that maps concept representations to steering interventions, trained end-to-end with an optimal transport loss. New concepts, including ones never seen at training time, are served at deployment by a single forward pass (0.08s on \nitro{}, 0.26s on \dmdtwo{}), eliminating the seconds-to-minutes per-concept optimization needed by standard steering methods.

\item \textbf{Zero-shot Cross-modal Conditioning.} To our knowledge, \method is the first activation-steering method to take images as conditioning signal: at deployment, a reference image can directly steer text-to-image generation with no retraining (\Cref{sec:exp_crossmodal}).

\item \textbf{Comprehensive Evaluation.} Across $167$ held-out concepts on \dmdtwo and \nitro, \methodshort matches the strongest per-concept baselines at applying the target concept while preserving the original generation nearly as well, at a fraction of their cost. At comparable inference cost to prompting, humans and VLMs (\Cref{tab:vlm_human_preference}) pick \method roughly $2{\times}$ as often as prompting.
\end{itemize}

\section{Related Work}
\label{sec:related}

Controllable generation has evolved from fine-tuning to activation steering, with recent work exploring hypernetwork-based approaches. We position \method within this progression.

\paragraph{Activation Steering.}
Activation steering modifies internal model representations to control generation without changing the weights of the model. ITI~\citep{li2024inference} uses classifier directions to steer toward truthfulness. CAA~\citep{rimsky2023steering} applies contrastive activation addition for behavior control. Linear-AcT~\citep{linearact} and LinEAS~\citep{lineas} use optimal transport to learn affine transformations between source and target activation distributions. Other works address toxicity mitigation~\citep{suau2024whispering} and reasoning understanding~\citep{venhoff2025understandingreasoningthinkinglanguage}. While effective, all these methods require \textit{per-concept optimization}, limiting scalability.

\paragraph{Hypernetwork-Based Adaptation.}
Hypernetworks~\citep{ha2016hypernetworks} generate weights of a target network conditioned on an input. Recent work applies them to controllable generation: Text2LoRA~\citep{charakorn2025texttolora} predicts LoRA~\citep{hulora} parameters from text, HyperLoRA~\citep{li2025hyperlora} targets portrait synthesis, and HyperDreamBooth~\citep{ruiz2024hyperdreambooth} predicts DreamBooth-imitation LoRA offsets for Text-to-Image (T2I) subject personalization but still requires ${\sim}20$s of per-subject fine-tuning, whereas \method{} is trained end-to-end through the generator and operates in activation space with no further optimization at deployment. HyperSteer~\citep{sun2025hypersteeractivationsteeringscale} is the closest prior work: it also trains a hypernetwork end-to-end to predict activation-steering parameters on the fly, sharing our goal of eliminating
per-concept optimization. \methodshort differs on four axes: (i) the steering mechanism is affine-elementwise optimal transport rather than additive CAA-style, inheriting the bounded strength parameter $\lambda \in [0, 1]$ of Linear-AcT/LinEAS; (ii) training uses an unpaired distributional loss (1D Wasserstein) rather than paired supervision; 
(iii) the target is a T2I generator rather than an LLM and 
(iv) the hypernetwork is a simple MLP ${\sim}1000\times$ smaller than the target model, whereas HyperSteer's is same-size (a constraint that is a practical limitation).

\paragraph{Cross-Model Transfer.}
A growing body of work demonstrates that neural networks develop structurally similar latent spaces across architectures~\citep{moschella2023relative,huh2024platonic}. This enables cross-model knowledge transfer: steering vectors can be transferred between models via linear mappings~\citep{huang-etal-2025-cross,bello2025linearrepresentationtransferabilityhypothesis}, and safety interventions can transfer across model families~\citep{oozeer2025activationspaceinterventionstransferred}. 
\method leverages this insight to learn a knowledge mapping from CLIP's structured embedding space to the activation space of the target model, effectively achieving complete inference-time decoupling from the expensive target model.

\section{Method}
\label{sec:method}

We present \method, a framework that learns a hypernetwork to predict interventions from arbitrary concept encodings. In this section, we first discuss concept representation (\Cref{sec:concepts}), then review standard activation steering (\Cref{sec:steering}), and finally introduce our approach for amortized intervention prediction (\Cref{sec:hypertransport}, \Cref{sec:architecture}).

\subsection{Concepts and Data}
\label{sec:concepts}

A concept $c$ is represented by a set of target sentences $\{\vx_c\}\sim\mathcal{X}_c$ that embody the concept. For instance, for the concept ``Van Gogh style'', target sentences might include ``\textit{Bold, expressive brushstrokes, vibrant colors, and swirling textures}''. 
To train a hypernetwork, we require a collection of concepts $\mathcal{C} = \{c_1, \ldots, c_K\}$, each with its associated target sentences $\mathcal{X}_{c_k}$. 

We focus on globally-applied visual concepts (styles, palettes, environmental settings; full list in \Cref{app:targetconcepts}), the regime in which the optimal-transport steering we amortize (\Cref{sec:steering}) has been shown to work well. Localized edits would require a different parameterization (pixel-local or cross-attention) and are outside the scope of this work.   

Additionally, we rely on a shared set of source sentences $\mathcal{X}^\src$ that do not contain any of the target concepts. Model activations over $\mathcal{X}^\src$ define the neutral distribution from which we steer away.

\subsection{Background: Activation Steering}
\label{sec:steering}

\paragraph{Intervention Framework.}
Given a target generative model $\Model$ and desired concept $c$, an intervention $\Interv_{\theta_c}$ modifies the internal activations of $\Model$ to steer generation toward the expression of $c$. Formally, the intervention produces steered activations $\actv_{c} = \Interv_{\theta_c}(\actv)$.
In practice, activations are grouped by layers in $\Model$ so that $\actv = \{\actv^{(1)},\ldots,\actv^{(\ell)},\ldots,\actv^{(L)}\}$, which are intervened on by layer-specific parameters $\theta_c = \{\theta_c^{(1)}, \ldots, \theta_c^{(\ell)}, \ldots, \theta_c^{(L)}\}$. In our notation, $L$ refers to the number of layers being intervened, not the total number of layers in $\Model$, which may not be the same.

\paragraph{Per-Concept Optimization.}
Common activation steering methods (re)optimize interventions independently for each concept, using a dataset to that end:
\begin{equation}
\label{eq:perconcept}
\theta_c^* = \argmin_{\theta}
\underset{
\scriptsize\begin{array}{r@{\,}c@{\,}l}
    \vx&\sim&\mathcal{X}^\src \\
    \vx_c&\sim&\mathcal{X}_c
\end{array}}{\E}
\Big[
\mathcal{L}_{\text{align}}\big(\Model(\vx; \theta), \Model(\vx_c)\big)
\Big],
\end{equation}
where $\mathcal{L}_{\text{align}}$ measures alignment between steered generations (using an intervention $\Interv_{\theta_c}$ and source prompts $\mathcal{X}^\src$) and generation from target prompts $\mathcal{X}_c$. Here, $\Model(\cdot)$ denotes any relevant output from $\Model$, \eg internal activations.

We focus on steering interventions of the affine elementwise form introduced by Linear-AcT~\citep{linearact}: $\Interv_{\theta_c}(\actv) = (1-\lambda)\actv + \lambda(\vw\odot\actv + \vb)$, with $\theta_c = [\vw, \vb]^\top$ and $\odot$ the elementwise product. The parameterization makes two scope choices. First, the same affine map is applied at every spatial location, suiting it to global, low-frequency properties (artistic styles, palettes, scene-wide attributes) rather than spatially localized edits. Second, the map is fitted once and applied during a single forward pass, suiting it to one-step generative models rather than multi-step denoising trajectories where activation distributions shift across steps. This is the operating regime of the per-concept methods we amortize, and our baselines (CAA, ITI, Linear-AcT, LinEAS) all live in it; \method holds the parameterization and intervention sites fixed across baselines, isolating amortization as the variable under study rather than parameterization design. Within this functional form, the baselines differ only in how $\theta_c$ is estimated. CAA~\citep{rimsky2023steering} uses contrastive activation addition over pairs, and ITI~\citep{li2024inference} uses the normal of a logistic classifier separating source and target activations; both reduce to additive interventions ($\vw = \mathbf{1}$). Linear-AcT~\citep{linearact} and LinEAS~\citep{lineas} estimate $\vw$ and $\vb$ in closed form by minimizing a sum of 1D optimal transport losses. While effective, all four require a \textit{separate optimization for each new concept}, with per-concept latency of seconds to minutes that rules out interactive deployment. With \method we amortize this per-concept estimation (\cref{eq:perconcept}): a single hypernetwork maps a concept representation to $\theta_c$ in one forward pass, for any concept including those unseen at training time.

\subsection{\method: Amortized Interventions}
\label{sec:hypertransport}

To amortize per-concept optimization costs, we introduce a trainable hypernetwork $\Hyper: \mathcal{E}\mapsto\Theta$ that predicts intervention parameters from concept encodings $\crep_c = \Enc(\vx_c)$. The concept encoding space $\mathcal{E}$ is produced by a pre-trained, frozen encoder $\Enc: \mathcal{X}\mapsto \mathcal{E}$ mapping data to embeddings. Intervention prediction is achieved by composing the encoder and hypernetwork: $\theta_c = \Hyper(\crep_c) = \Hyper \circ \Enc(\vx_c)$.

\paragraph{End-to-End Training.}
We train $\Hyper$ by directly optimizing the alignment objective, applying predicted interventions on $\Model$ during training:
\begin{equation}
\label{eq:hypertransport}
\Hyperpsi^* = \argmin_\Hyperpsi\hspace{-7mm}
\underset{
\scriptsize\begin{array}{r@{\,}c@{\,}l}
    c&\sim&\mathcal{C} \\
    \vx&\sim&\mathcal{X}^\src \\
    \vx_c,\vx_c^\prime&\sim&\mathcal{X}_c
\end{array}}{\E}
\hspace{-5mm}
\Big[ \mathcal{L}_{\text{align}}\big(\Model(\vx; \Hyper \circ \Enc(\vx_c^\prime)), \Model(\vx_c)\big) \Big].
\end{equation}
This end-to-end formulation requires the alignment loss $\mathcal{L}_{\text{align}}$ to be differentiable. Following LinEAS~\citep{lineas}, we adopt the 1D Wasserstein distance as alignment objective. For each layer $\ell$ with $d_\ell$ neurons, the loss decomposes independently per neuron, noted $W$ in \Cref{fig:fig1}:

\begin{equation}
\label{eq:alignment_loss}
\mathcal{L}_{\text{align}} = \sum_{\ell=1}^{L} \sum_{j=1}^{d_\ell} W_p\!\left(\hat{a}_j^{(\ell)},\; a_{c,j}^{(\ell)}\right),
\end{equation}
where $\hat{a}_j^{(\ell)}$ denotes the steered activation produced by applying the predicted intervention $\theta_c=\Hyper\circ\Enc(\vx_c^\prime)$ at neuron $j$ of layer $\ell$, and $a_{c,j}^{(\ell)}$ denotes the corresponding target activation. For equal-sized distributions, the $p$-Wasserstein distance on a single activation dimension reduces to
\begin{equation}
\label{eq:wp_closed_form}
W_p(\hat{\va}, \va) = \left(\frac{1}{n}\sum_{i=1}^{n}\left|\hat{a}_{(i)} - a_{(i)}\right|^p\right)^{1/p},
\end{equation}
where $\hat{a}_{(i)}, a_{(i)}$ are the order statistics of the steered and target activations. This requires only sorting (a permutation, so gradients flow to the underlying activations) and an element-wise computation, so no iterative optimal transport solver is needed. While LinEAS uses $p{=}2$, we found $p{=}1$ to perform better empirically and use it throughout.
This optimal transport loss provides a principled, well-behaved objective that yields state-of-the-art steering performance. The name \method reflects hypernetwork-based prediction combined with this transport-based alignment objective.

\paragraph{Amortized Inference and Flexibility} 
When a new steering intervention is required, even for a concept that was not seen during training, the user can provide one or more sentences that describe the desired concept. When more than one sentence is provided, we average their embeddings (encoded by $\Enc$) and provide the result as input to the hypernetwork $\Hyper$ that in a single inference pass produces all the steering weights for the target model. Since this avoids the per-concept optimization on $\Model$ that LinEAS and similar methods require, the marginal cost per concept drops from hundreds of seconds to a fraction of a second (see \Cref{tab:main_comparison} and \Cref{app:percost}).

Crucially, \method requires $\mathcal{E}$ to be structured enough for $\Hyper$ to generalize to unseen conditions. Arbitrary encodings such as one-hot could in principle be learned, but they would not support generalization.
In practice, we use CLIP~\citep{radford2021clipg}: its contrastive vision-language pretraining aligns text and images at the level of image-wide attributes, matching our concept set. The shared text/image space also lets us condition on images instead of text (\cref{sec:exp_crossmodal}). \Cref{app:encoder_robustness} compares several encoders on this task and finds CLIP-based variants generalize best to unseen concepts; \method{} is agnostic to this choice and can be retrained with any encoder.

\subsection{\method Architecture}
\label{sec:architecture}

For practical deployment, $\Hyper$ should be far smaller than $\Model$ in terms of parameters and FLOPS required to compute the intervention weights. Additionally, $\Hyper$ should be flexible enough to predict intervention parameters at various layers of $\Model$, which may have different dimensionalities and statistics. We opt for simplicity and implement $\Hyper$ as an MLP that directly accepts inputs in $\mathcal{E}$ (full architecture in \Cref{app:mlp}). In practice, the inputs are the average encodings of target sentences $\mathcal{X}_c$, and the MLP directly outputs the intervention parameters, in our case $\theta_c =[\vw,\vb]^\top$. We also experimented with a more expressive \perceiver~\citep{jaegle2022perceiver} variant that handles per-sentence encodings without averaging; we report it in \Cref{app:perceiver} along with training-curve comparisons (\Cref{app:training_curves}), and find that the added complexity does not yield meaningful gains in our low-data regime (\ie 32 sentences for each concept, with the hypernetwork predicting interventions concept-wise).

\section{Experiments}
\label{sec:experiments}

\subsection{Experimental Setup}
\label{sec:exp_setup}
\paragraph{Dataset.}\label{par:dataset}
We curate a dataset of 777 visually rich concepts through systematic LLM prompting across 10 conceptual categories, with a minimum per-category of 44 concepts (``digital methods'') and maximum of 124 (``environmental settings''). For each concept, we prompt a frontier model to generate $32$ representative image prompts with the explicit instruction of being suitable for text-to-image generation.
We apply CLIPScore~\citep{hessel2021clipscore} filtering to ensure semantic consistency and distinctiveness. Details and examples of the generated sentences are provided in \Cref{app:dataset}, and an analysis of the train/test concept distribution in \Cref{app:concept_distribution}.

\paragraph{Models and Baselines.}
As target generative model $\Model$ we experiment with \textbf{\dmdtwo} and \textbf{\nitro}. Both are single-step (distilled) diffusion models, a setting we inherit from \lineas which fits activation steering research, as they do not require timestep conditioning and enable efficient forward passes during optimization.
To test the effectiveness of \method, we compare against four per-concept activation steering methods: \textbf{CAA}, \textbf{ITI}, \textbf{Linear-AcT} and \textbf{LinEAS}.
All steering baselines share identical experimental conditions: the same intervention positions, element-wise affine parameterization, training data, and intervention strength ($\lambda{=}1$); only the estimation procedure differs across methods. We intervene after all normalization layers in the ResNet part of \dmdtwo (layers matching \texttt{unet.*norm.*}) and the transformer layers of \nitro (\texttt{transformer.transformer\_blocks.[0-9]+.norm[12]}). Per-concept baselines use an incremental estimation variant that is strictly stronger than their original formulations (\Cref{app:baselines}). \methodshort is trained with the $p{=}1$ alignment loss of \Cref{eq:wp_closed_form} (ablation in \Cref{app:l1_l2}).
We also report \textbf{Prompting} as a reference: a mechanistically different conditioning, included for its wide adoption, its practical utility, and the increasing controllability it offers as models scale~\citep{cheng-genctrl}.

\begin{table*}[tb]
\centering
\small
\caption{\textbf{\method matches \lineas on 167 held-out concepts} (best \conceptfidelity on \nitro; within $0.02$ on \promptfidelity for both models), \textbf{at the cost of a single hypernetwork forward pass per new concept.} Held-out here means unseen by \method, while CAA, ITI, Linear-AcT, and \lineas are trained from scratch on each test concept, and \textit{Time} is their per-concept training cost. For \method, \textit{Time} ($^\star$) is the inference cost of producing intervention weights from a concept encoding. End-to-end training of \method on the 610 training concepts takes ${\sim}64$ GPU-hours (8 GPUs $\times$ 8h), against ${\sim}170$ GPU-hours to fit \lineas individually on the same set. Prompting is included as a cross-paradigm reference (\Cref{sec:exp_prompting}); training-concept results in \Cref{tab:main_comparison_full}; per-concept inference gap in \Cref{app:percost}.}
\label{tab:main_comparison}
\setlength{\tabcolsep}{4pt} 
\begin{tabular}{lcccccc}
\toprule
 & \multicolumn{3}{c}{\dmdtwo} & \multicolumn{3}{c}{\nitro} \\
\cmidrule(lr){2-4} \cmidrule(lr){5-7}
Method & Time (s) $\downarrow$ & \promptfidelityshort $\uparrow$ & \conceptfidelityshort $\uparrow$
       & Time (s) $\downarrow$ & \promptfidelityshort $\uparrow$ & \conceptfidelityshort $\uparrow$ \\
\midrule
Unsteered      & --              & $0.467$                       & $0.204 \pm 0.025$
               & --              & $0.440$                       & $0.213 \pm 0.029$              \\
\midrule
CAA            & $641$           & $0.285 \pm 0.048$             & $0.275 \pm 0.041$
               & $138$           & $0.334 \pm 0.056$             & $0.254 \pm 0.038$              \\
ITI            & $1290$          & $0.305 \pm 0.043$             & $0.275 \pm 0.040$
               & $320$           & $0.279 \pm 0.060$             & $0.254 \pm 0.040$              \\
LinAcT         & $775$           & $0.355 \pm 0.045$             & $\mathbf{0.322 \pm 0.046}$
               & $167$           & $0.391 \pm 0.024$             & $\underline{0.265 \pm 0.040}$  \\
LinEAS         & $936$           & $\underline{0.431 \pm 0.027}$ & $0.278 \pm 0.044$
               & $543$           & $\underline{0.398 \pm 0.023}$ & $0.264 \pm 0.039$              \\
\midrule
Prompting      & --              & $\mathbf{0.446 \pm 0.016}$    & $0.262 \pm 0.036$
               & --              & $\mathbf{0.416 \pm 0.014}$    & $0.261 \pm 0.036$              \\
\midrule
HT (ours)      & $0.258^\star$   & $0.416 \pm 0.026$             & $\underline{0.280 \pm 0.036}$
               & $0.078^\star$   & $0.390 \pm 0.020$             & $\mathbf{0.266 \pm 0.038}$     \\
\bottomrule
\end{tabular}
\end{table*}

\begin{table*}[tb]
\centering
\small
\caption{\textbf{\method vs.\ Prompting: VLM and human evaluation.} Two-stage evaluation on 167 held-out test concepts. Stage~1 assesses binary \promptfidelity (images judged individually) and only images that pass it for both methods go to the next one. Stage~2 reports the head-to-head share of \conceptfidelity preferences, restricted to ratings that picked one of the two methods (\textsc{Both}/\textsc{None} excluded). Human study: 19 participants, $135$ of $285$ Stage-2 responses expressed a preference. Bold marks the per-column winner. Un-normalized breakdown including \textsc{Both}/\textsc{None} responses in \Cref{tab:vlm_human_preference_full}; VLM details in \Cref{app:vlm_study}.}
\label{tab:vlm_human_preference}
\setlength{\tabcolsep}{6pt}
\begin{tabular}{llcccc}
\toprule
& & \multicolumn{2}{c}{\textit{Stage 1:} \promptfidelity} & \multicolumn{2}{c}{\textit{Stage 2:} \conceptfidelity} \\
& & \multicolumn{2}{c}{\scriptsize (\% faithful)} & \multicolumn{2}{c}{\scriptsize (\% preferred, head-to-head)} \\
\cmidrule(lr){3-4} \cmidrule(lr){5-6}
\textit{Model} & & \textit{VLM} & \textit{Human} & \textit{VLM} & \textit{Human} \\
\midrule
\multirow{2}{*}{\dmdtwo}
  & Prompting & $\mathbf{83.1}$ & $\mathbf{57.7}$ & $30.5$          & $29.7$          \\
  & \method   & $70.1$          & $43.7$          & $\mathbf{69.5}$ & $\mathbf{70.3}$ \\
\midrule
\multirow{2}{*}{\nitro}
  & Prompting & $\mathbf{82.7}$ & $\mathbf{51.7}$ & $37.0$          & $29.6$          \\
  & \method   & $71.1$          & $39.9$          & $\mathbf{63.0}$ & $\mathbf{70.4}$ \\
\bottomrule
\end{tabular}
\end{table*}

\paragraph{Metrics.}
We assess interventions along two axes: how effectively they induce the desired concept in the generated image, and how well they preserve the content expressed in the original prompt. CLIPScore~\citep{hessel2021clipscore} is the standard tool for both: it follows the evaluation protocol of related per-concept steering work~\citep{linearact,lineas} and of the broader text-to-image generation and concept-control literature~\citep{gandikota2023concept,betker2023improving}. We adopt it here because (i) it is cheap enough to run across our full evaluation grid (167 test concepts $\times$ 100 prompts $\times$ multiple methods), and (ii) it gives a strong relative signal when compared across methods that share the same intervention form and prompts, which is the setting throughout \Cref{sec:exp_main} 
Concretely, let $\vy = \Model(\vx; \theta_c)$ be the image generated from source prompt $\vx$ under intervention $\Interv_{\theta_c}$, and $t_c$ the textual representation of concept $c$. We then define $\promptfidelity = \mathrm{CLIPScore}(\vy, \vx)$, measuring preservation of the original prompt, and $\conceptfidelity = \mathrm{CLIPScore}(\vy, t_c)$, measuring induction of the target concept.
We additionally measure the time required to obtain a new intervention. For traditional per-concept steering methods, this is the time required to \textit{train} the intervention. For \method, the training time is amortized since it allows to generate interventions for new concepts without retraining. The cost of generating such an intervention is the \textit{inference} time on $\Hyper \circ \Enc(\vx_c)$.
\subsection{Comparison Against Existing Conditioning Paradigms}
\label{sec:exp_compare}

We train \method on $610$ of the $777$ concepts in our dataset (an $\sim$80/20 split; \cref{par:dataset}), each represented by 32 sentences. At inference, for fair comparison with the other methods, interventions are produced by conditioning on all available target sentences; we analyze sensitivity to this choice in \Cref{sec:exp_oneshot}. We evaluate all methods on the 167 test concepts never seen during \methodshort training; per-concept baselines are optimized from scratch on each test concept. \Cref{tab:main_comparison} reports the results, analyzed below against per-concept steering (\Cref{sec:exp_main}) and against prompting (\Cref{sec:exp_prompting}).

\subsubsection{\method vs.\ Per-Concept Baselines}
\label{sec:exp_main}

\begin{figure}[tb]
    \centering
    \includegraphics[width=\linewidth]{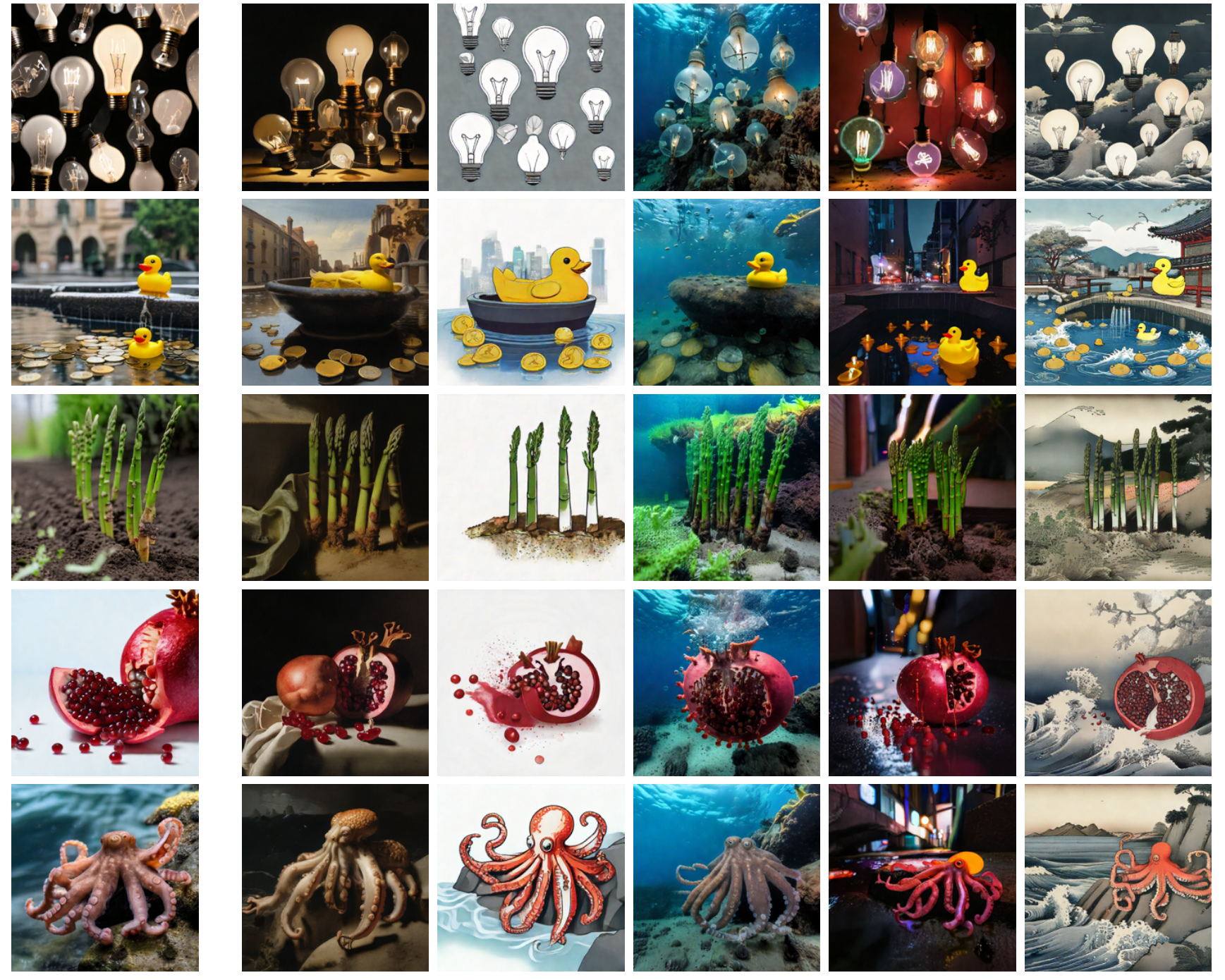}
    \caption{\textbf{\textbf{Test} concept conditioned generations using \method} on \dmdtwo. On the left we show images generated without steering. Other columns show generations obtained via \methodshort steering, for concepts (left to right) \{\textit{Caravaggio, Vector Graphics, Underwater, Neon alley, Ukiyo-e}\}.\label{fig:grid}}
\end{figure}

Among steering methods, \method{} operates as a high-speed approximation of LinEAS: it trades a small \promptfidelity decrement (within $0.02$ on both models: $0.416$ vs.\ $0.431$ on \dmdtwo; $0.390$ vs.\ $0.398$ on \nitro) for matched \conceptfidelity ($0.280$ vs.\ $0.278$ on \dmdtwo; $0.266$ vs.\ $0.264$ on \nitro) at $3600$--$7000\times$ lower per-concept latency. At deployment, a new intervention is produced in $0.258$s on \dmdtwo and $0.078$s on \nitro through a single hypernetwork pass, against the $936$s and $543$s respectively that LinEAS requires to fit each concept from scratch. The total-cost comparison over the full training set is reported in \Cref{tab:main_comparison}; \Cref{app:percost} contextualizes the per-concept inference gap.  

In \Cref{fig:grid} we show qualitative examples of \method steerings with test concepts, showing the effectiveness of our approach at predicting steerings \textit{never} seen during training. The leftmost image is unconditional (no steering applied to the model $\Model$), while the remaining columns show steerings for the five test concepts named in \Cref{fig:grid}. Rows relate to different prompts (top to bottom):
\textit{{``Light bulbs of different shapes piled together with some broken", 
``Rubber duck floating in a city fountain among the coins.", 
``Asparagus spears pushing up through the soil in a garden row.", 
``A pomegranate split open with seeds spilling onto the surface.", 
``An octopus stretching its tentacles across tide pool rocks"}}. More qualitative examples in \Cref{fig:qualitative_comparison} and in the appendix (\Cref{app:qualitative}).

\subsubsection{\method vs.\ Prompting}
\label{sec:exp_prompting}

Within the steering paradigm (\Cref{sec:exp_main}), all baselines share the same intervention form and are evaluated on matched images, so CLIP-based metrics serve as a fair relative ranking.
Here, prompting and steering produce qualitatively different outputs, and CLIP-based metrics are known to misalign with human perception in T2I~\citep{otani2023toward}. We therefore complement the CLIP metrics with two judge protocols (human and VLM) that decouple \promptfidelity from \conceptfidelity.

\paragraph{User Study.}\label{sec:userstudy}
We conducted a comparative user study with human judges to assess preference between images generated using prompting alone, and those steered with \method. To decouple \promptfidelity from \conceptfidelity, the study is organized in two intentionally separated stages. \textbf{Stage 1 (Prompt Faithfulness):} each judge rates $30$ individual images for faithfulness to the original prompt (\textsc{Yes} / \textsc{No}), ignoring the applied artistic style. \textbf{Stage 2 (Concept Comparison):} the judge sees $15$ side-by-side image pairs generated from the same prompt by the two methods, together with the concept name, definition, and up to three indicative reference images, and selects which image better displays the concept (\textsc{Image 1} / \textsc{Image 2} / \textsc{Both} / \textsc{None}). The study involved 19 human judges, yielding $n{=}285$ Stage-2 responses; in $135$ of these, the judge preferred one of the two images (Image~1 or Image~2), and in the remaining $150$ they selected \textsc{Both} or \textsc{None}. Exact instructions and a screenshot of the interface used in the study are in~\cref{app:userstudy}.

When human judges expressed a preference between the two images, they chose \method $95$ out of $135$ times ($70.4\%$), a $2{:}1$-or-better preference that holds for both generators (\Cref{tab:vlm_human_preference}). Un-normalized breakdown including \textsc{Both}/\textsc{None} responses in \Cref{tab:vlm_human_preference_full}.

\paragraph{VLM-as-a-Judge.}
To complement the user study at scale, we conduct a VLM-as-a-judge evaluation using \vlmmodel~\citep{chen2024internvl} on test image pairs, following the same two-stage protocol as above (full prompts in \Cref{app:vlm_study}). Among pairs where both methods produce prompt-faithful outputs and the VLM judge expressed a preference, \method is preferred head-to-head $69.5\%$ on \dmdtwo and $63.0\%$ on \nitro (\Cref{tab:vlm_human_preference}), corroborating the user-study findings.

\subsection{Cross-Modal Conditioning}
\label{sec:exp_crossmodal}

Since \method conditions on CLIP embeddings, it naturally supports cross-modal inputs: images can steer text-to-image generation without retraining. This is the second capability \method offers that prompting cannot. Neither natural-language prompts nor traditional steering methods can take an image as a conditioning signal.
We evaluate image-conditioned steering where users provide reference images rather than text descriptions. CLIP encodes reference images, and \method predicts interventions from these embeddings. Importantly, \method was trained only on text conditioning.
\Cref{tab:cross_modal_results} shows that image conditioning achieves comparable performance to text conditioning, demonstrating zero-shot cross-modal transfer. Under the VLM-as-judge protocol used above, image-conditioned \method is preferred $39.6\%$ vs.\ $23.5\%$ for prompting on prompt-faithful pairs (\Cref{app:vlm_study}), closely tracking the text-conditioned result in \Cref{tab:vlm_human_preference} and confirming that cross-modal transfer preserves concept articulation quality.

\begin{table}[t]
\centering
\small
\caption{\textbf{Zero-shot Cross-modal Steering} performance on test concepts using \dmdtwo with 32 conditionings. We combine different training modalities with conditioning ones, obtaining zero-shot settings when they are mismatched. More fine-grained results for DMD2 in \Cref{app:results}.}
\label{tab:cross_modal_results}
\begin{tabular}{lcccc}
\toprule
\textbf{Train Modality} & \multicolumn{2}{c}{Text} & \multicolumn{2}{c}{Image} \\
\cmidrule(lr){2-3} \cmidrule(lr){4-5}
\textbf{Condition Modality} & Text & Image & Image & Text \\
\midrule
\promptfidelity $\uparrow$  & $0.416 \pm 0.026$ & $0.402 \pm 0.035$ & $0.414 \pm 0.031$ & $0.237 \pm 0.038$ \\
\conceptfidelity $\uparrow$ & $0.280 \pm 0.036$ & $0.283 \pm 0.037$ & $0.274 \pm 0.038$ & $0.351 \pm 0.046$ \\
\bottomrule
\end{tabular}
\end{table}

\begin{figure}[t]
\centering
    \includegraphics[width=0.7\linewidth]{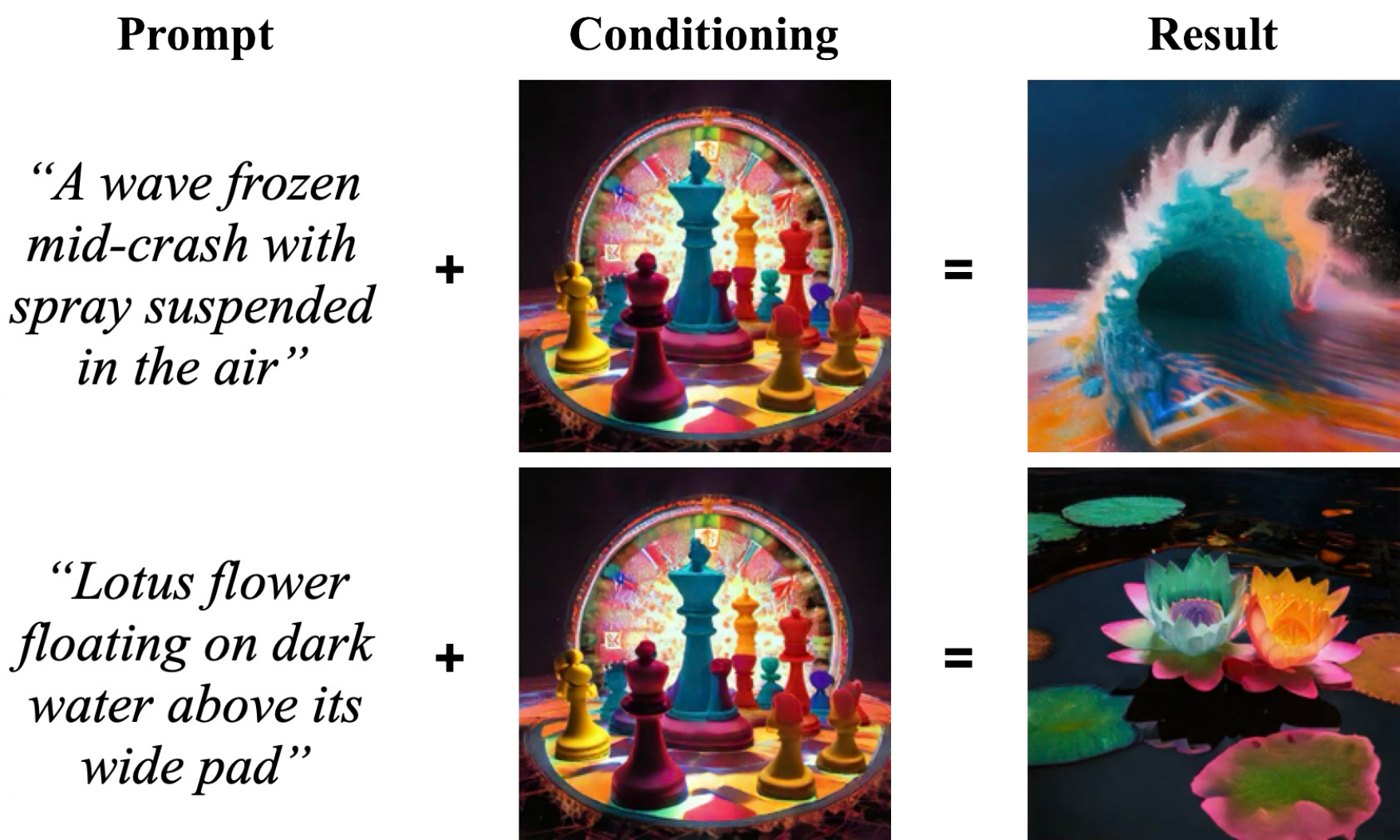}
\caption{Cross-modal control: images can condition text-to-image steering without retraining. From left to right: the prompt, the image used as conditioning for the hypernetwork to produce an intervention, the steered generation.
}
\label{fig:cross_modal_examples}
\end{figure}

\textbf{Qualitative Examples.} \Cref{fig:cross_modal_examples} shows steering results across conditioning modalities. Image-based conditioning successfully captures stylistic attributes from reference images while being faithful to the original prompt. We remark that, even when conditioning on images, the time required to instantiate a steering with \method is similar to that of text conditioning.

\section{Conclusion}
\label{sec:conclusion}

\method maps a frozen encoder embedding to the affine-elementwise intervention parameters that per-concept methods optimize from scratch, at prompting-like deployment cost. Empirically, it matches \lineas at applying the target concept at $3600$--$7000\times$ lower per-concept latency, and is preferred over prompting by both human and VLM judges at roughly $2{:}1$. Decoupling concept representation from intervention prediction yields three properties no prior approach combines: cost-amortized steering for unseen concepts, bounded strength control $\lambda \in [0, 1]$ from optimal-transport steering, and cross-modal conditioning. \method is most useful when the concept set is large, evolving, or specified at request time; per-concept methods remain competitive on small fixed sets, and prompting suffices when text alone elicits the target.

\paragraph{Limitations.} \method inherits the two scope choices of the affine elementwise parameterization~\citep{lineas,linearact}, shared by all per-concept baselines (CAA, ITI, Linear-AcT, LinEAS): they are limitations of the underlying steering family, not of our amortization. First, the same affine map governs every spatial location, suiting global, low-frequency properties (artistic styles, palettes, scene-wide attributes) but not spatially localized edits. Second, it is fitted once and applied in a single forward pass, suiting one-step generative models (\dmdtwo, \nitro) but not multi-step denoising trajectories whose activation distributions shift across steps. Either extension calls for a different functional form, which our hypernetwork framework can amortize given additional conditioning signals (\eg the timestep for multi-step diffusion).
The encoder defines the concept coverage and intervention space applicability, and should be chosen to match the target domain. To this end, our formulation is encoder-agnostic: replacing CLIP with a domain-specific or stronger encoder only requires retraining the hypernetwork.
Following prior work, we use a single strength $\lambda$ across all normalization layers; choosing a subset of layers or letting $\lambda$ vary per layer could improve efficiency and quality.

\section{Broader Impact}
\label{sec:broader_impact}

\method targets the cost and brittleness of concept-level conditioning of generative models. We outline below the potential positive and negative societal impacts of \methodshort and the mitigations our framework affords.

\paragraph{Positive impacts.}
By amortizing per-concept optimization into a single forward pass (\Cref{app:percost}), \method lowers the compute and engineering overhead of steering frozen generators. We see three concrete benefits.
(i) \emph{Accessibility:} concept-level control becomes affordable for users and labs that cannot run a per-concept fitting pipeline at request time, broadening who can adapt open-weight generative models.
(ii) \emph{Safety-relevant conditioning:} the same machinery that steers a model \emph{toward} a target concept can steer it \emph{away} from one, giving practitioners a lightweight handle for content control on top of existing model-level safeguards. (iii) \emph{Auditable knobs:} the bounded strength parameter $\lambda \in [0,1]$ from optimal-transport steering~\citep{lineas} (\Cref{app:control}) makes the magnitude of the intervention explicit and graded, which is easier to log, threshold, and review than free-form prompt edits.

\paragraph{Negative impacts and intended-use harms.}
\method shares the misuse surface of any concept-steering method applied to a pretrained generator. We highlight two categories that follow directly from the technology working as intended. (i) \emph{Style and persona impersonation:} cheap, request-time conditioning on a target concept can be used to mimic the style of a specific artist or the appearance of a specific public figure without consent. The base generator's existing safeguards remain the primary line of defense, but \method does not add new ones. (ii) \emph{Amplification of biased or unsafe content:} if the concept set used to train or query \method contains harmful, stereotyping, or unsafe descriptions, the hypernetwork will faithfully amortize them. Because \method is encoder-agnostic, biases in the underlying CLIP embedding (or any replacement encoder) are inherited by the predicted interventions.

\paragraph{Harms from incorrect operation.}
When \method underfits a concept, the failure mode is benign: the generator falls back toward its unconditional behavior, which is the same distribution a user would have sampled without steering. When \method overfits or distorts a concept, the resulting samples are visibly off-distribution rather than convincingly deceptive, which limits (but does not eliminate) the risk of weaponizing failure cases for disinformation.

\paragraph{Mitigations.}
Several properties of the method support mitigation rather than requiring external machinery. The bounded $\lambda \in [0,1]$ control surface lets deployers cap intervention strength globally or per concept. Training data is generated entirely from LLM prompts (\Cref{app:dataset}); curating, filtering, or red-teaming this concept set is a direct lever on what \method will be able to steer toward at deployment. Because \method does not modify the base generator's weights, all model-level safeguards (safety filters, watermarking, identity blocklists) of the underlying generator continue to apply unchanged. We do not release a new generative model or scraped dataset, so the additional misuse surface introduced by this work is bounded by the artifacts of the base models we build on.

\clearpage
\bibliographystyle{abbrvnat}
\bibliography{bibliography}

\appendix

\clearpage
\FloatBarrier
\section{Dataset Details}
\label{app:dataset}

\begin{table}[h]
  \centering
  \begin{tabular}{lc}
  \toprule
  \textbf{Category} & \textbf{Number of Concepts} \\
  \midrule
  Environmental Settings & 124 \\
  Art Techniques & 100 \\
  Historical Periods & 99 \\
  Color Treatments & 80 \\
  Photography \& Cinema & 80 \\
  Fantasy Genres & 77 \\
  Illustration Styles & 64 \\
  Individual Artists & 61 \\
  Artistic Movements & 47 \\
  Digital Methods & 44 \\
  Generic ($\mathcal{X}^\src$) & 1 \\
  \midrule
  \textbf{Total} & \textbf{777} \\
  \bottomrule
  \end{tabular}
  \caption{Distribution of concepts across categories. The singleton ``Generic'' category serves as the general source one; min/max counts cited in the main body refer to the ten thematic categories.}
  \label{tab:style-categories}
  \end{table}

\begin{figure}[h]
  \centering
  \includegraphics[width=0.8\textwidth]{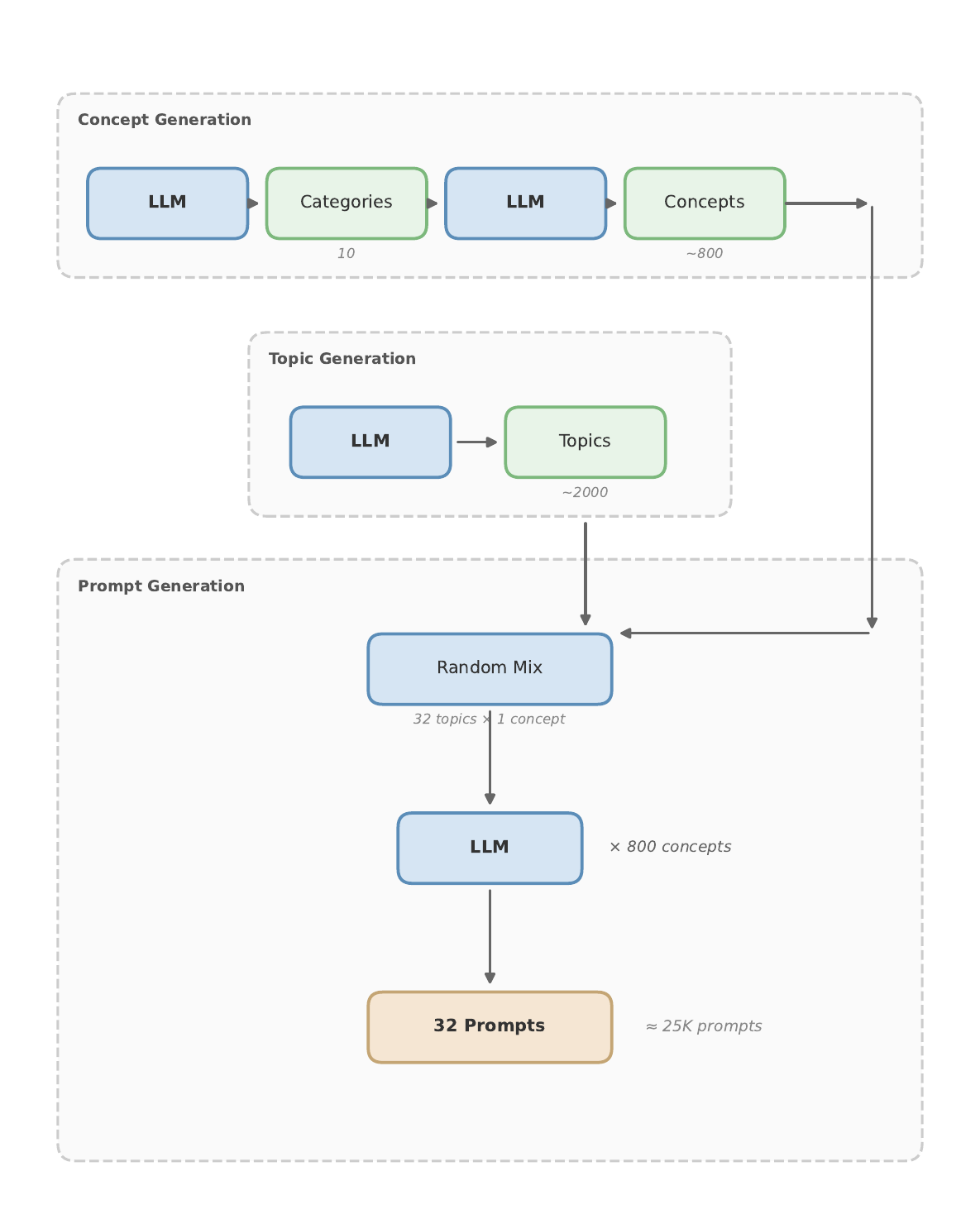}
  \caption{\textbf{Prompt generation pipeline.} Starting from 10 thematic categories elicited from a frontier LLM, we expand each into a set of visually-rich concepts (${\sim}800$ in total) and, in a parallel pass, build a pool of ${\sim}2000$ generic topics. For every concept we then sample 32 topics uniformly and prompt the LLM to recombine each (concept, topic) pair into a single image-generation prompt, yielding ${\sim}25\text{K}$ prompts.}
  \label{fig:dataset_pipeline}
\end{figure}

\FloatBarrier
\clearpage
\section{Implementation Details: MLP Architecture}

\label{app:mlp}

\begin{figure}[h]
  \centering
  \includegraphics[width=0.6\textwidth]{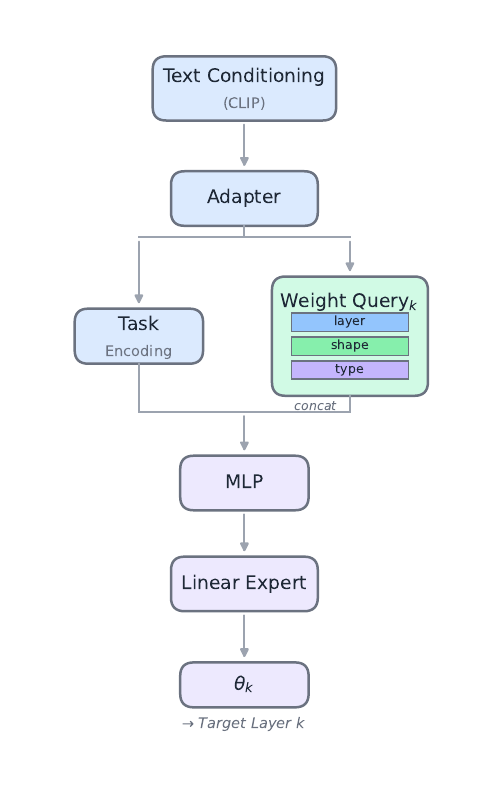}
  \caption{Architecture of the MLP-based hypernetwork. Our MLP architecture processes CLIP embeddings through multiple layers to predict intervention parameters for each layer of the target model.}
  \label{fig:mlp_architecture}
\end{figure}

\begin{table}[h]
\centering
\caption{Hypernetwork architecture breakdown by target model. Rows above the parameter breakdown are configuration; rows below are parameter counts. Layer-embedding composable settings (\texttt{key}=128, \texttt{shape}=64, \texttt{state\_key}=64, \texttt{output\_projection}=True) are shared across both targets.}
\label{tab:app:hyper_breakdown}
\begin{tabular}{lrr}
\toprule
 & \dmdtwo & \nitro \\
\midrule
Adapter output dim        & 256       & 256 \\
Weight query dim          & 256       & 256 \\
Task+weight query dim     & 512       & 512 \\
Intervened modules        & 256 (LN+GN, U-Net) & 56 (LN, Transformer) \\
Weight queries            & 512       & 112 \\
Max output $\theta_k$     & 2{,}560   & 1{,}152 \\
\midrule
\multicolumn{3}{l}{\emph{Parameter breakdown}} \\
Input adapters            & 197{,}120     & 197{,}120 \\
Layer embeddings          & 131{,}584     & 80{,}064 \\
Decoding projection (MLP) & 3{,}674{,}112 & 3{,}674{,}112 \\
Single linear expert      & 1{,}313{,}792 & 591{,}488 \\
\midrule
Total                     & 5{,}904{,}768 & 4{,}671{,}808 \\
\bottomrule
\end{tabular}
\end{table}

\paragraph{Training.}
We train $\Hyper$ end-to-end with AdamW~\citep{loshchilov2019decoupled} at learning rate $10^{-4}$ (default $\beta_1{=}0.9$, $\beta_2{=}0.999$, $\epsilon{=}10^{-8}$) under a cosine schedule that dampens the learning rate by a factor of $10^{-3}$ at the end of training, maintaining an exponential moving average (EMA) of the hypernetwork weights with decay $0.99$ and using the EMA copy at evaluation. The hypernetwork is kept in $\mathrm{float32}$; the encoder $\Enc$ and the target generator $\Model$ are kept frozen (the target generator runs in $\mathrm{bfloat16}$). One epoch corresponds to a single pass over the $610$ training concepts in random order; for each concept we sample a fresh subset of its sentences as conditioning input $\vx_c^\prime$ and as target-distributi
on samples $\vx_c$, so the specific sentences seen for a given concept vary across epochs. Training is data-parallel: concepts are sharded across $8$ H100 GPUs (per-rank batch of $1$ concept), with gradients synchronized via standard all-reduce; the full $300$-epoch run takes ${\sim}8$h ($\sim$64 GPU-hours total, see \Cref{tab:main_comparison} caption).

\FloatBarrier

\section{\perceiver Architecture.}
\label{app:perceiver}

Although our main results use a simpler MLP-based architecture, we also explored implementing $\Hyper$ with a \perceiver~\citep{jaegle2022perceiver} based one, which naturally accommodates heterogeneous inputs and outputs. The added complexity yielded no meaningful gains (-1 \conceptfidelity, +1 \promptfidelity), consistent with the low-data regime of our setting (32 samples per concept/class), where the more data-hungry Perceiver IO is at a disadvantage. We therefore default to the MLP, but describe the Perceiver-based variant below for completeness, as it may prove beneficial in settings with richer or more heterogeneous data.

The architecture comprises three stages:

\paragraph{Cross-attention encoding.} Processes an arbitrary number of concept encodings $\crep_c$, enabling flexible conditioning on single or multiple examples.

\paragraph{Self-attention processing.} Refines latent representations while conditioning on the concept encodings. The number and dimensionality of latent representations is flexible.

\paragraph{Specialized decoders.} Generate layer-specific interventions $\{\theta_c^{(\ell)}\}$ conditioned on refined latent representations and learnable intervention queries for each target layer.

This architecture cleanly separates concept understanding (encoding) from intervention generation (decoding), allowing the hypernetwork to generalize across concepts and target layers.

The Perceiver-based hypernetwork uses:
\begin{itemize}
    \item 256 latent tokens with dimension 512
    \item 4 cross-attention layers for encoding
    \item 6 self-attention layers for processing
    \item Separate decoder heads for each target layer
\end{itemize}

\clearpage
\section{Baseline Implementation}
\label{app:baselines}

Our implementation of \textbf{CAA}, \textbf{ITI}, and \textbf{Linear-AcT} is strictly stronger than their original layer-independent formulations: we estimate interventions \emph{incrementally}, layer by layer in forward order, with each layer's objective solved on \emph{already-intervened} activations from prior layers. This accounts for inter-layer dependencies the original formulations ignore: a correction applied at layer $\ell{-}1$ shifts the activation distribution observed at layer $\ell$, so fitting layer $\ell$ to the perturbed distribution yields a tighter solution than fitting it to the unperturbed one. All baselines share the same intervention positions (\Cref{sec:exp_setup}) and the same strength $\lambda{=}1$.

\section{Additional Results}
\label{app:results}

\begin{figure}[h]
\centering
    \includegraphics[width=0.5\linewidth]{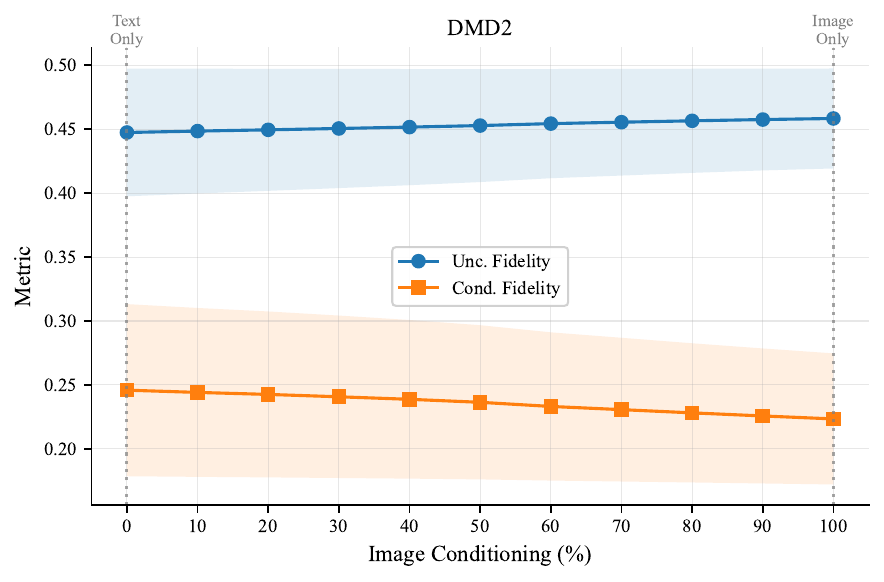}
\caption{Varying percentage of image conditioning over 32 total for \method}
\label{fig:app:multimodal}
\end{figure}

\subsection{Robustness to Encoding Space}
\label{app:encoder_robustness}
We compare \method trained with several pretrained encoders to identify which embedding space is best suited for predicting interventions. We test multimodal encoders (CLIP variants and the SDXL~\citep{podellsdxl} input encoder) against text-only sentence encoders: BGE~\citep{bge_embedding}, ModernBERT~\citep{modernbert}, MPNet\footnote{\myurl{https://huggingface.co/sentence-transformers/all-mpnet-base-v2}}, and MiniLM\footnote{\myurl{https://huggingface.co/sentence-transformers/all-MiniLM-L6-v2}}. Each hypernetwork is trained for 300 epochs under identical conditions.

\Cref{tab:encoder_robustness} reports the comparison. On training concepts, every encoder except ModernBERT performs comparably (\conceptfidelity from 0.282 for text-only to 0.288 for CLIP variants; ModernBERT trails at 0.243). On test concepts, CLIP-based encoders are the strongest: their \conceptfidelity drops only ${\sim}0.02$ (from 0.288 to 0.268--0.271), while text-only encoders fall further (MiniLM goes from 0.282 to 0.251, an 11\% reduction). The takeaway is that CLIP's multimodal alignment yields the most structured embedding space for visual concepts, making it the best encoder choice for predicting interventions in this setting. ModernBERT shows the highest \promptfidelity but the weakest \conceptfidelity: it preserves prompts but does not induce concepts as effectively.

Concepts poorly represented in the chosen encoder's embedding space (\eg highly technical or domain-specific imagery) may yield less effective interventions. Integrating a domain-specific encoder requires only retraining the hypernetwork, with no architectural changes.

\begin{table*}[t]
\caption{\textbf{Robustness to encoder choice.} Performance of \method with different pretrained encoders, same evaluation setting as \Cref{tab:main_comparison}. Each hypernetwork is trained for 300 epochs to steer \dmdtwo. CLIP-based encoders achieve the best \conceptfidelity, with stronger generalization on test concepts.}
\label{tab:encoder_robustness}
\small
\centering
\begin{tabular}{lcccc}
  \toprule
  & \multicolumn{2}{c}{\textbf{Train}} & \multicolumn{2}{c}{\textbf{Test}} \\
  \cmidrule(lr){2-3} \cmidrule(lr){4-5}
  \textbf{Encoder} & \promptfidelityshort & \conceptfidelityshort & \promptfidelityshort & \conceptfidelityshort \\
  \midrule
  \multicolumn{5}{l}{\textit{Multimodal (CLIP-based)}} \\
  SDXL Input      & $0.426 \pm 0.032$ & $0.288 \pm 0.047$ & $0.435 \pm 0.021$ & $0.271 \pm 0.037$ \\
  CLIP ViT-bigG   & $0.426 \pm 0.032$ & $0.288 \pm 0.047$ & $0.436 \pm 0.020$ & $0.270 \pm 0.036$ \\
  CLIP ViT-L      & $0.427 \pm 0.032$ & $0.287 \pm 0.047$ & $0.438 \pm 0.020$ & $0.269 \pm 0.035$ \\
  CLIP ViT-B      & $0.427 \pm 0.032$ & $0.287 \pm 0.047$ & $0.438 \pm 0.019$ & $0.268 \pm 0.036$ \\
  \midrule
  \multicolumn{5}{l}{\textit{Text-only}} \\
  BGE Large        & $0.429 \pm 0.032$ & $0.283 \pm 0.046$ & $0.442 \pm 0.017$ & $0.258 \pm 0.034$ \\
  MPNet            & $0.429 \pm 0.031$ & $0.282 \pm 0.046$ & $0.443 \pm 0.017$ & $0.255 \pm 0.035$ \\
  MiniLM           & $0.430 \pm 0.031$ & $0.282 \pm 0.046$ & $0.444 \pm 0.016$ & $0.251 \pm 0.035$ \\
  ModernBERT Large & $0.449 \pm 0.017$ & $0.243 \pm 0.037$ & $0.454 \pm 0.009$ & $0.230 \pm 0.029$ \\
  ModernBERT Base  & $0.449 \pm 0.019$ & $0.243 \pm 0.035$ & $0.454 \pm 0.010$ & $0.231 \pm 0.028$ \\
  \bottomrule
\end{tabular}

\end{table*}

\subsection{One-Shot Capability}
\label{sec:exp_oneshot}

\begin{table*}[h]
\caption{\textbf{One-shot comparison on \dmdtwo.} At $N{=}1$, per-concept baselines exhibit distinct failure modes: LinEAS overfits (high \conceptfidelity but degraded \promptfidelity), Linear-AcT retreats to a conservative intervention, and CAA/ITI approach the unsteered baseline. \method degrades gracefully, maintaining competitive performance on both metrics.}
\label{tab:one_shot}
\scriptsize
\centering
\setlength{\tabcolsep}{2pt}
\begin{tabularx}{0.95\textwidth}{l*{2}{>{\centering\arraybackslash}X}}
  \toprule
   \textbf{Method} & \promptfidelity $\uparrow$ & \conceptfidelity $\uparrow$ \\
  \midrule
    \textit{Unsteered} & $0.467$ & $0.205 \pm 0.025$ \\
   \cmidrule(lr){1-3}
    LinEAS & $0.328 \pm 0.058$ & $0.307 \pm 0.050$ \\
    Linear-AcT & $0.358 \pm 0.044$ & $0.296 \pm 0.041$ \\
    CAA & $0.451 \pm 0.013$ & $0.215 \pm 0.029$ \\
    ITI & $0.459 \pm 0.007$ & $0.207 \pm 0.026$ \\
   \cmidrule(lr){1-3}
    \method & $0.441 \pm 0.016$ & $0.251 \pm 0.034$ \\
  \bottomrule
\end{tabularx}
\end{table*}

\begin{figure}[h]
\centering
    \includegraphics[width=0.5\linewidth, trim=0 0 0 20, clip]{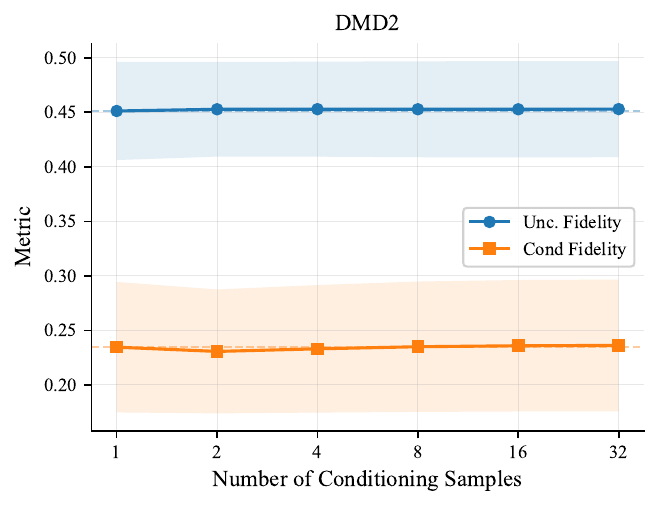}
\caption{One/Few-shot \method showing good performance at all regimes, from $N=1$ (one-shot) to $N=32$.}
\label{fig:fewshot}
\end{figure}

A practical advantage of \method is its effectiveness with minimal conditioning data. While in our main comparison (\Cref{tab:main_comparison}) we condition on 32 target sentences that describe each concept, real-world deployment benefits from simpler conditioning requirements. For example, users may only have a single example or brief description of their desired concept.
In \Cref{tab:one_shot} we compare the performance at $N=1$ versus that at $N=32$, where $N=1$ represents the most user-friendly one-shot scenario (see \Cref{fig:fewshot} for intermediate values $N=2,4,8,16$).
We observe that reducing the conditioning samples from 32 to 1 increases the fidelity to the prompt (\promptfidelityshort) at the expense of a moderate reduction in conditioning capability ($\sim$10\% drop in \conceptfidelityshort). This trade-off is expected: with fewer conditioning examples, the model has less information to constrain the concept representation, allowing it to remain more faithful to the original prompt while slightly reducing its ability to capture the full nuances of the target concept. Despite this expected degradation, the results show that \method can instantiate new steerings, even on unseen test concepts, with access to only one sentence, which is not achievable with any of the traditional steering methods in the literature.

\begin{table*}[htb]
\centering
\small
\caption{Full results on training and test concepts. Training concepts are those used during \method amortization; test concepts are held out. Per-concept baselines are trained individually on each concept in both settings.}
\label{tab:main_comparison_full}
\begin{tabular}{clcccc}
\toprule
\multicolumn{2}{c}{} & \multicolumn{2}{c}{\textbf{Train}} & \multicolumn{2}{c}{\textbf{Test}} \\
\cmidrule(lr){3-4} \cmidrule(lr){5-6}
 \textit{Model} & \textit{Method} & \promptfidelityshort & \conceptfidelityshort & \promptfidelityshort & \conceptfidelityshort \\
\midrule
\multirow{7}{*}{\begin{sideways}DMD2\end{sideways}}
& \textit{Unsteered} & \textit{$0.467 \pm 0.000$} & \textit{$0.204 \pm 0.023$} & \textit{$0.467 \pm 0.000$} & \textit{$0.204 \pm 0.025$} \\
\cmidrule(lr){2-6}
& CAA & $0.286 \pm 0.046$ & $0.278 \pm 0.042$ & $0.285 \pm 0.048$ & $0.275 \pm 0.041$ \\
& ITI & $0.307 \pm 0.046$ & $0.279 \pm 0.042$ & $0.305 \pm 0.043$ & $0.275 \pm 0.040$ \\
& Linear-AcT & $0.358 \pm 0.043$ & $\mathbf{0.321 \pm 0.047}$ & $0.355 \pm 0.045$ & $\mathbf{0.322 \pm 0.046}$ \\
& LinEAS & $\underline{0.432 \pm 0.024}$ & $0.280 \pm 0.043$ & $\underline{0.431 \pm 0.027}$ & $0.278 \pm 0.044$ \\
\cmidrule(lr){2-6}
& Prompting & $\mathbf{0.447 \pm 0.014}$ & $0.261 \pm 0.037$ & $\mathbf{0.446 \pm 0.016}$ & $0.262 \pm 0.036$ \\
\cmidrule(lr){2-6}
& \method (ours) & $0.404 \pm 0.040$ & $\underline{0.295 \pm 0.046}$ & $0.416 \pm 0.026$ & $\underline{0.280 \pm 0.036}$ \\
\midrule[0.5pt]
\multirow{7}{*}{\begin{sideways}Nitro-1-PixArt\end{sideways}}
& \textit{Unsteered} & \textit{$0.440 \pm 0.000$} & \textit{$0.212 \pm 0.025$} & \textit{$0.440 \pm 0.000$} & \textit{$0.213 \pm 0.029$} \\
\cmidrule(lr){2-6}
& CAA & $0.338 \pm 0.053$ & $0.258 \pm 0.036$ & $0.334 \pm 0.056$ & $0.254 \pm 0.038$ \\
& ITI & $0.280 \pm 0.057$ & $0.256 \pm 0.042$ & $0.279 \pm 0.060$ & $0.254 \pm 0.040$ \\
& Linear-AcT & $0.391 \pm 0.024$ & $\underline{0.267 \pm 0.037}$ & $0.391 \pm 0.024$ & $\underline{0.265 \pm 0.040}$ \\
& LinEAS & $\underline{0.399 \pm 0.023}$ & $\underline{0.267 \pm 0.042}$ & $\underline{0.398 \pm 0.023}$ & $0.264 \pm 0.039$ \\
\cmidrule(lr){2-6}
& Prompting & $\mathbf{0.417 \pm 0.015}$ & $0.262 \pm 0.038$ & $\mathbf{0.416 \pm 0.014}$ & $0.261 \pm 0.036$ \\
\cmidrule(lr){2-6}
& \method (ours) & $0.387 \pm 0.025$ & $\mathbf{0.274 \pm 0.041}$ & $0.390 \pm 0.020$ & $\mathbf{0.266 \pm 0.038}$ \\
\bottomrule
\end{tabular}
\end{table*}

\subsection{Loss Norm Ablation: $p{=}1$ vs.\ $p{=}2$}\label{app:l1_l2}
We compare \method trained with $p{=}1$ (L1) vs.\ $p{=}2$ (L2) in the Wasserstein alignment loss of \Cref{eq:alignment_loss}, evaluated on the same 167 held-out test concepts as \Cref{tab:main_comparison}.

\begin{table}[h]
\centering
\small
\caption{\textbf{L1 vs.\ L2 alignment loss.} \method trained with $p{=}1$ vs.\ $p{=}2$ in \Cref{eq:wp_closed_form}, evaluated on the 167 held-out test concepts of \Cref{tab:main_comparison} for both \dmdtwo and \nitro.}
\label{tab:l1_l2_ablation}
\begin{tabular}{llcc}
\toprule
\textit{Model} & \textit{Loss} & \promptfidelityshort $\uparrow$ & \conceptfidelityshort $\uparrow$ \\
\midrule
\multirow{2}{*}{\dmdtwo}
  & $p{=}1$ (L1) & $0.416 \pm 0.055$ & $0.280 \pm 0.057$ \\
  & $p{=}2$ (L2) & $0.437 \pm 0.020$ & $0.271 \pm 0.036$ \\
\midrule
\multirow{2}{*}{\nitro}
  & $p{=}1$ (L1) & $0.391 \pm 0.064$ & $0.266 \pm 0.054$ \\
  & $p{=}2$ (L2) & $0.402 \pm 0.016$ & $0.255 \pm 0.034$ \\
\bottomrule
\end{tabular}
\end{table}

\FloatBarrier
\clearpage
\section{Per-concept inference cost}
\label{app:percost}

\Cref{tab:main_comparison} reports two distinct cost quantities: the per-concept training cost of each baseline (CAA, ITI, Linear-AcT, LinEAS) and \method's per-concept inference cost. For per-concept baselines, this is the time to fit a new intervention via optimization on the target model: $936$s on \dmdtwo and $543$s on \nitro for LinEAS, the strongest baseline. For \method, this is the time of a single hypernetwork forward pass: $0.258$s on \dmdtwo and $0.078$s on \nitro. The nominal ratios are ${\sim}3600\times$ and ${\sim}7000\times$ respectively.

These two quantities are not directly comparable: \method's inference cost follows a one-off training phase (${\sim}64$ GPU-hours over 610 concepts; see \Cref{tab:main_comparison} caption), whereas the baselines' time is incurred at every new concept. The per-concept inference comparison is informative when novel interventions must be produced at deployment, e.g.\ under interactive latency constraints or when the concept vocabulary is large, evolving, or unknown ahead of time.

\FloatBarrier
\clearpage
\section{User Study}
\label{app:userstudy}
In this section we provide more information about the user study described in \cref{sec:userstudy}: the exact instructions given to participants and a screenshot of the interface.

The study is organized in two stages that are intentionally separated so that judgments on \promptfidelity do not bias judgments on \conceptfidelity: a single-image \emph{prompt faithfulness} stage, followed by a pairwise \emph{concept comparison} stage. The same two-stage protocol is later re-used, in adapted form, in the VLM-as-a-judge experiment (\cref{app:vlm_study}).

The instructions provided to each user were as follows:

\begin{tcolorbox}[colback=gray!10, colframe=black, title=User Study Instructions]

This study has \textbf{TWO stages}. Please complete them in order.

\medskip
\textbf{Stage 1: Prompt Faithfulness}

In this stage you will see 30 images, one at a time.
Each image was generated from a text prompt using a specific  style (concept).

\medskip
\textbf{Your task:} Judge whether the image is \textbf{faithful} to the prompt.
\begin{itemize}
    \item Focus on the semantic content: are the objects, actions, relationships, and counts correct?
    \item \textbf{Ignore the artistic style} — the image has been intentionally stylized.
    \item Respond with \textbf{YES} (faithful) or \textbf{NO} (not faithful).
\end{itemize}

Please judge each image independently.

\medskip
\textbf{Stage 2: Concept Comparison}

In this stage you will see 15 pairs of images side by side.
Both images in each pair were generated from the \textbf{same prompt} and should display the \textbf{same concept/style}.
They were generated by two different methods.

\medskip
For each pair you will see:
\begin{itemize}
    \item The concept name and its definition
    \item Up to 3 example images showing what the concept looks like
    \item \textbf{Note:} These examples are purely indicative and should not be taken as a strong reference.
    They were found automatically and some of them might not be representative of the concept.
    \item Two generated images (Image 1 and Image 2)
\end{itemize}

\medskip
\textbf{Your task:} Which image better displays the concept?
\begin{itemize}
    \item \textbf{Image 1}: if Image 1 better displays the concept
    \item \textbf{Image 2}: if Image 2 better displays the concept
    \item \textbf{BOTH}: if both images display the concept equally well
    \item \textbf{NONE}: if neither image displays the concept well
\end{itemize}

\end{tcolorbox}

\begin{figure*}
    \centering
    \includegraphics[width=0.85\linewidth]{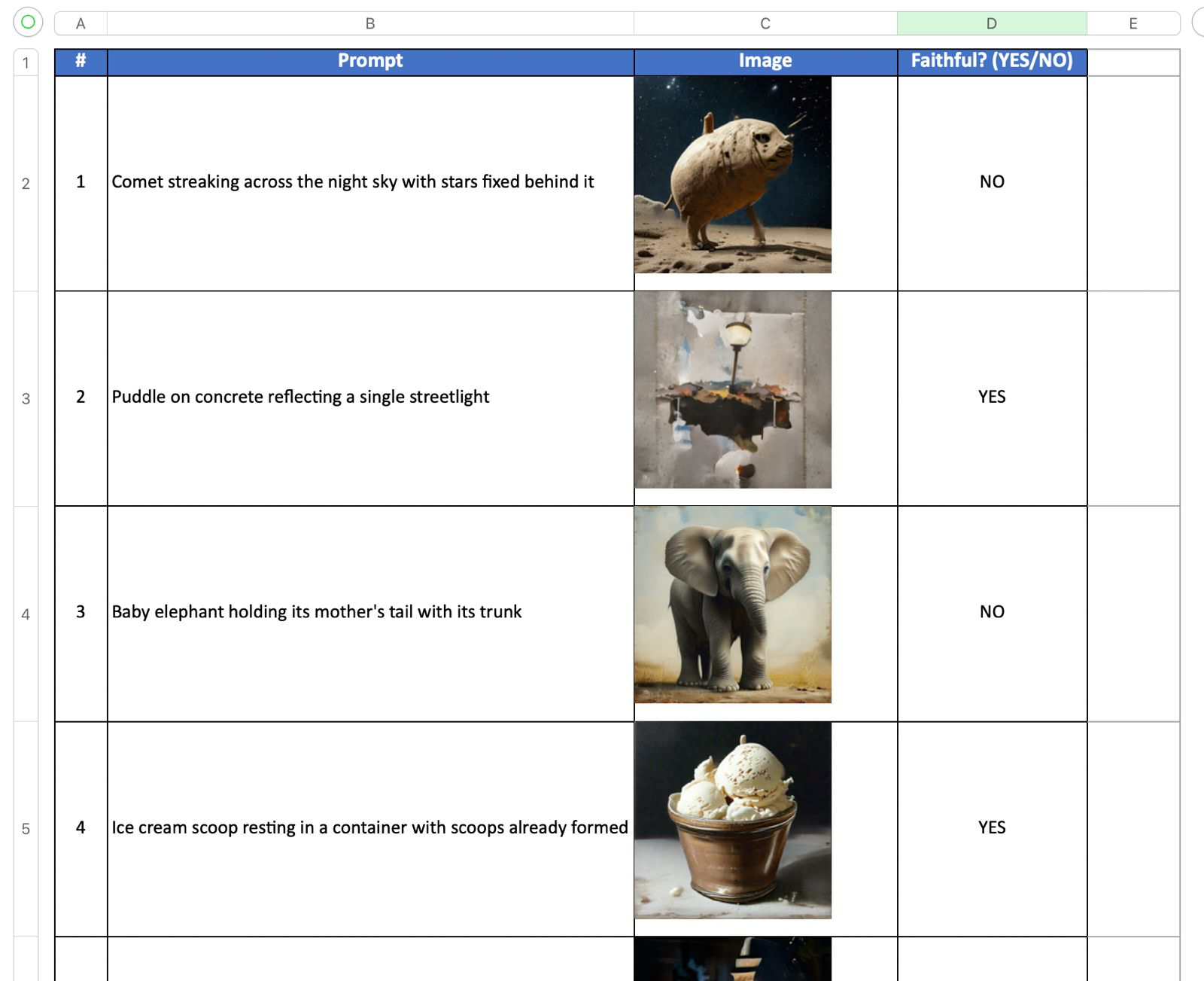}\\[0.6em]
    \includegraphics[width=0.95\linewidth]{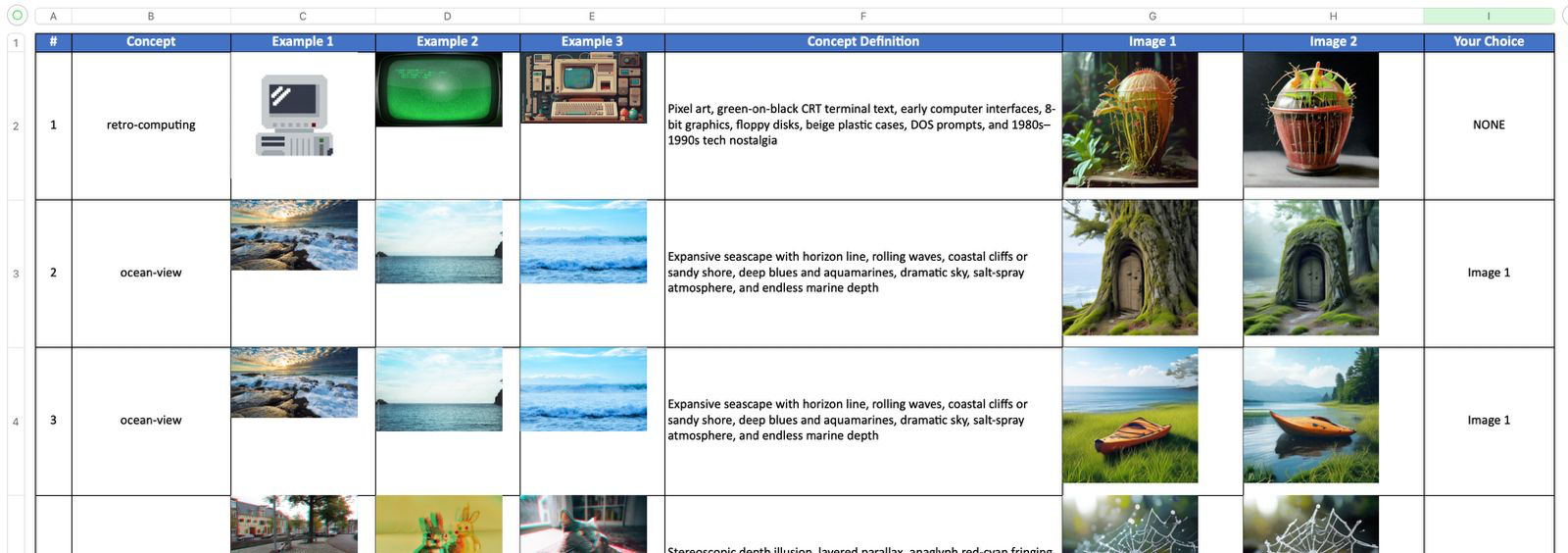}
    \caption{Examples of the user-study UI. \textbf{Top: Stage~1 (Prompt Faithfulness).} Each row shows the prompt and a generated image; the participant marks YES or NO for prompt-faithfulness. \textbf{Bottom: Stage~2 (Concept Comparison).} Each row shows the concept name, three indicative example images, the concept definition, and two generated images (Image~1 and Image~2, presented in random order from prompting or \method); the participant picks Image~1, Image~2, BOTH or NONE.}
    \label{fig:userstudyUI}
\end{figure*}

An example of how each request was shown to human judges is provided in \cref{fig:userstudyUI}. The choice of T2I model for each pair was randomized, with the constraint that the study contained an approximately equal number of generations from each model: in total $450$ pairs, $221$ with \dmdtwo and $229$ with \nitro. Each judge rated $15$ pairs sampled from this pool, yielding $n{=}285$ Stage~2 responses across 19 human judges; in $135$ of these, the judge picked one of the two images (\textsc{Image 1} or \textsc{Image 2}), and in the remaining $150$ they selected \textsc{Both} or \textsc{None}.

\begin{figure}[H]
    \centering
    \includegraphics[width=0.95\linewidth]{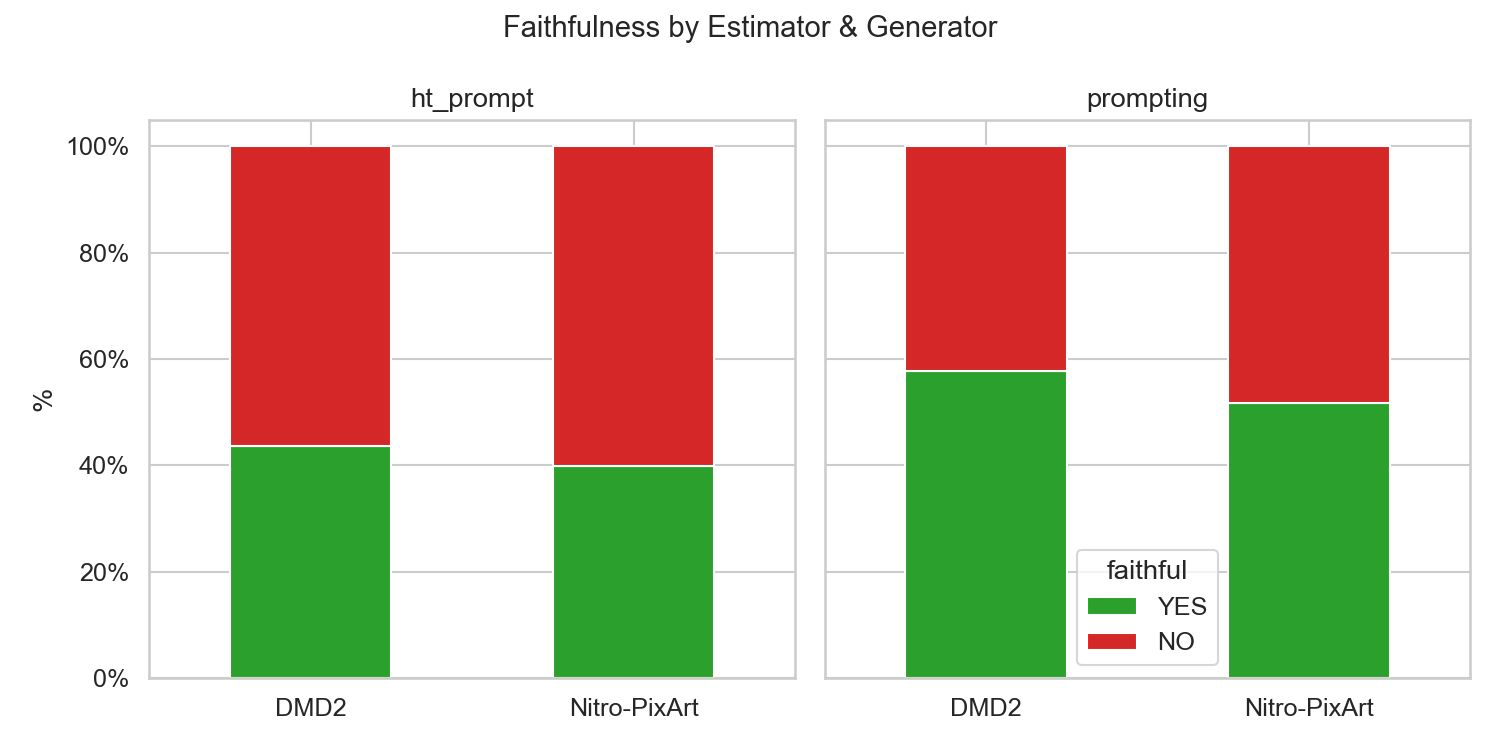}\\[0.6em]
    \includegraphics[width=0.95\linewidth]{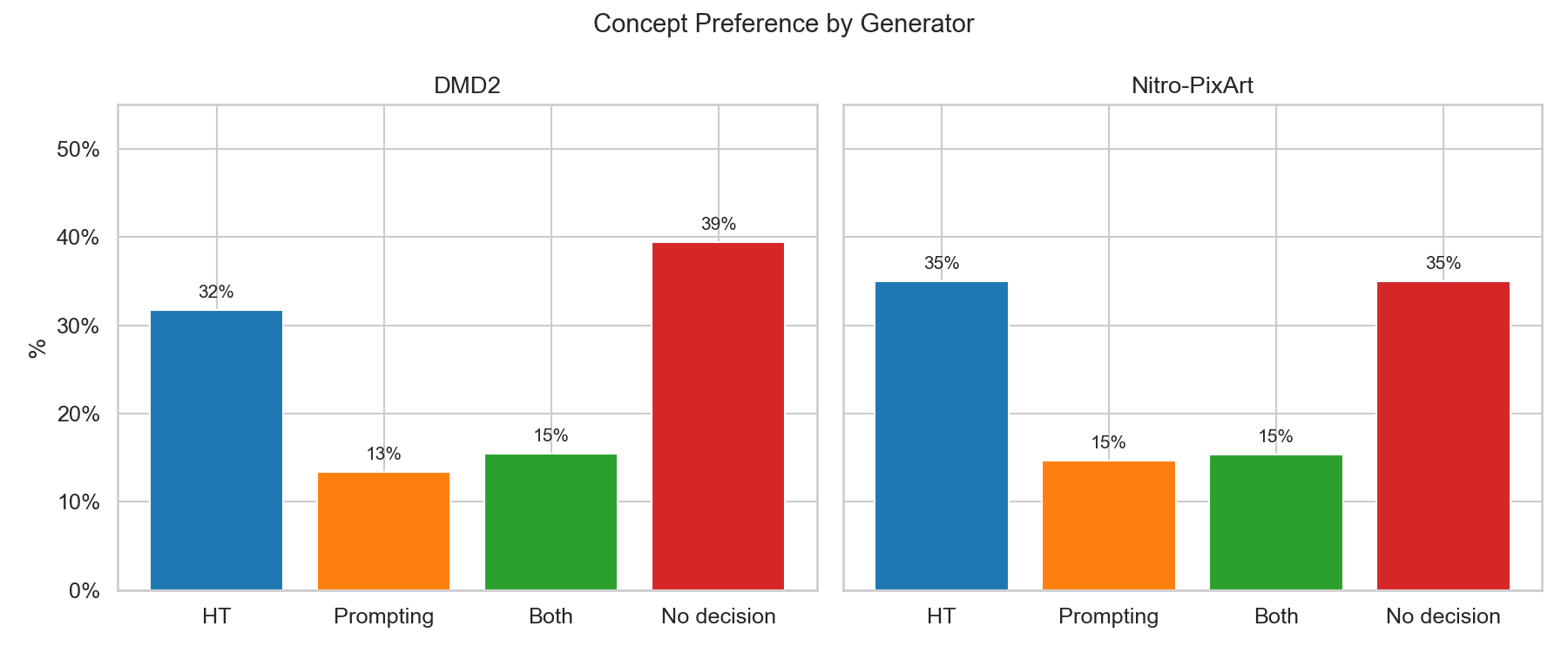}
    \caption{\textbf{Two-stage user study of \method generations} ($n{=}285$ Stage-2 responses from 19 human judges). \textbf{Top: Stage~1, Input Fidelity (per generator).} Prompting is slightly more often judged faithful than \method on both generators (aggregate $54.7\%$ \textit{vs.}\ $41.8\%$; McNemar $p{=}0.0001$, Cohen's $h{=}{-}0.261$, a small effect). \textbf{Bottom: Stage~2, Concept Fidelity (per generator).} Distribution of the four response options across all 285 pairs. Restricted to the 135 pairs where the judge picked one of the two images, \method is chosen $70.4\%$ of the time ($95/135$; $p<0.0001$, 95\% CI  $[63.0\%, 77.8\%]$, OR ${=}2.38$); per generator, $70.3\%$ on \dmdtwo and $70.4\%$ on \nitro (Fisher interaction $p{=}1.0$).}
    \label{fig:human_eval}
\end{figure}

\begin{figure}[H]
    \centering
    \includegraphics[width=\linewidth]{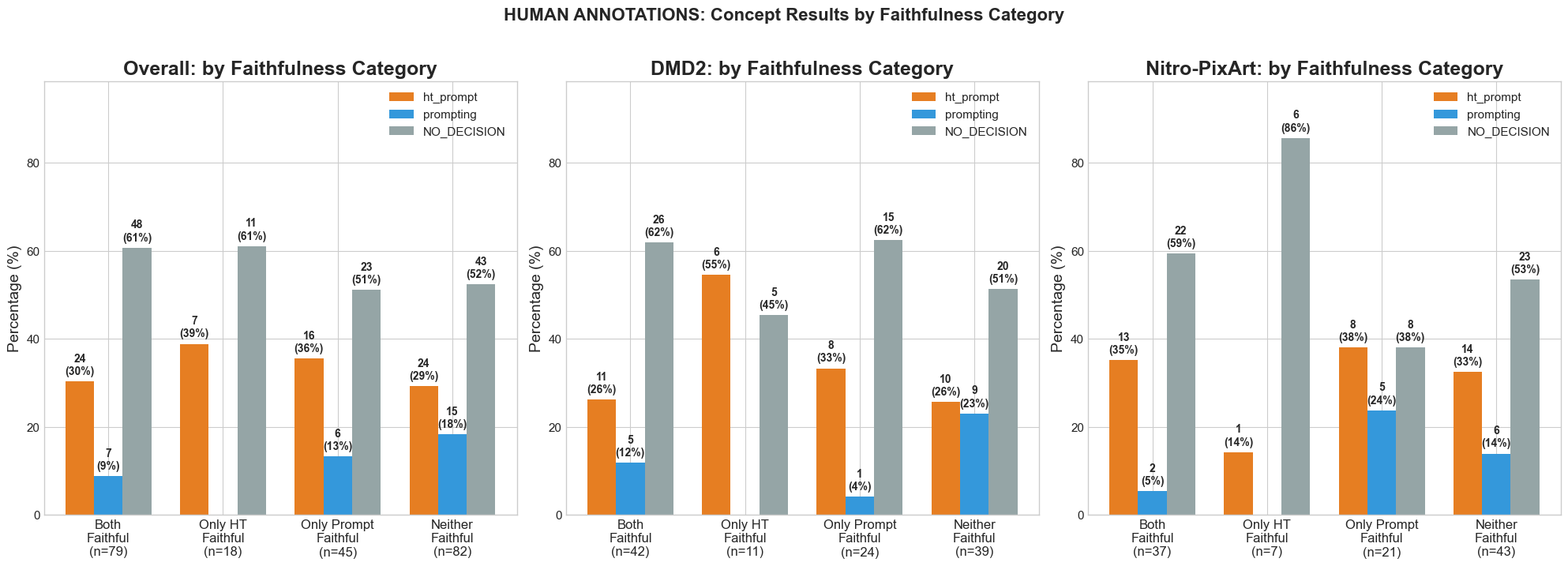}
    \caption{\textbf{Stage~2 (concept-fidelity) broken down by Stage~1 (input-fidelity) outcomes} (human annotations). \method is preferred across all faithfulness categories, including pairs where only the \emph{prompting} image was Stage-1 faithful (\method $36.7\%$ vs.\ prompting $16.7\%$). This is consistent with \method's 
    advantage being independent of input fidelity. ``No Decision'' rates remain high across categories.}
    \label{fig:human_eval_by_category}
\end{figure}

\begin{figure}[H]
    \centering
    \includegraphics[width=0.7\linewidth]{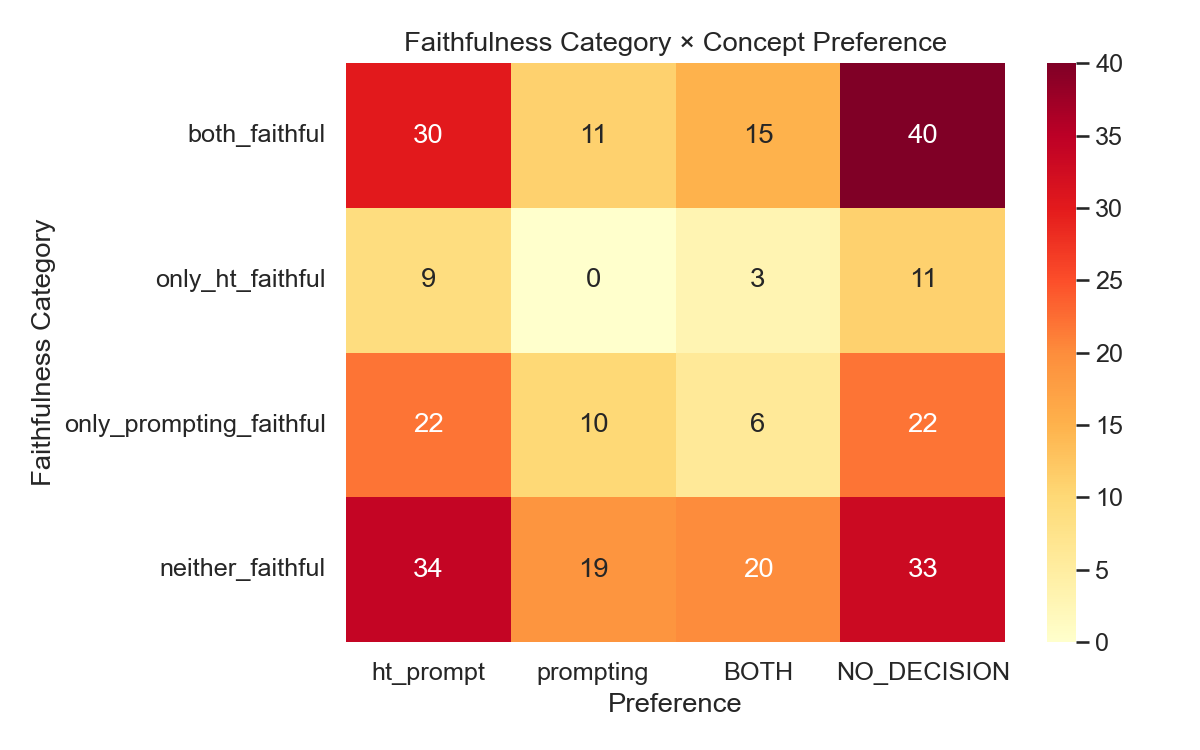}
    \caption{\textbf{Joint distribution of Stage-1 faithfulness category $\times$ Stage-2 response} (raw counts; rows are Stage-1 categories, columns are Stage-2 responses). When the human judge picks one of the two images (\emph{ht\_prompt} or \emph{prompting} columns), \method is chosen more often than prompting in every row, including \emph{only prompting faithful} ($22$ \textit{vs.}\ $10$): \method's concept-fidelity advantage holds even when only the prompting image is judged input-faithful. \textsc{Both} and \textsc{No Decision} are reported per row but excluded from the head-to-head share.}    \label{fig:concept_pref_heatmap}
\end{figure}

\paragraph{Robustness analyses.} The $70.4\%$ preference rate for \method is robust to alternative tie handling: scoring \textsc{Both} as half-credit to each method yields $65.4\%$, and scoring all ties (\textsc{Both} and \textsc{None}) as half-credit yields $59.6\%$, both well above the $50\%$ null. Across the $135$ pairs in which the human judge picked one of the two images, the winner is judged Stage-1 faithful in $44.4\%$ of cases and the loser in $46.7\%$ (essentially equal), inconsistent with a faithfulness-driven explanation. The cross-generator interaction is non-significant (Fisher $p{=}1.0$): \method's preference rate is $70.3\%$ on \dmdtwo and $70.4\%$ on \nitro.

\begin{table*}[tb]
\centering
\small
\caption{\textbf{Full breakdown of VLM and human evaluation, including \textsc{Both}/\textsc{None} responses.} Companion to \Cref{tab:vlm_human_preference}: same Stage-1 numbers, but Stage-2 here is \emph{un-normalized} and reports the raw share of each response category (Prompting / \method / \textit{No preference}), so each Stage-2 column sums to $100$ within a model. ``\textit{No preference}'' aggregates \textsc{Both} and \textsc{None} responses (Human) and unmatched/non-faithful pairs (VLM). For VLM, Stage~2 is over both-faithful pairs ($n_{\dmdtwo}{=}7827$, $n_{\nitro}{=}7377$); for Human, over all $285$ Stage-2 responses ($150$ of which were \textsc{Both}/\textsc{None}).}
\label{tab:vlm_human_preference_full}
\setlength{\tabcolsep}{6pt}
\begin{tabular}{llcccc}
\toprule
& & \multicolumn{2}{c}{\textit{Stage 1:} \promptfidelity} & \multicolumn{2}{c}{\textit{Stage 2:} \conceptfidelity} \\
& & \multicolumn{2}{c}{\scriptsize (\% faithful)} & \multicolumn{2}{c}{\scriptsize (\% of responses)} \\
\cmidrule(lr){3-4} \cmidrule(lr){5-6}
\textit{Model} & & \textit{VLM} & \textit{Human} & \textit{VLM} & \textit{Human} \\
\midrule
\multirow{3}{*}{\dmdtwo}
  & Prompting              & $\mathbf{83.1}$ & $\mathbf{57.7}$ & $20.5$          & $13.4$          \\
  & \method                & $70.1$          & $43.7$          & $\mathbf{46.7}$ & $\mathbf{31.7}$ \\
\cmidrule(lr){2-6}
  & \textit{No preference} & --              & --              & $32.8$          & $54.9$          \\
\midrule
\multirow{3}{*}{\nitro}
  & Prompting              & $\mathbf{82.7}$ & $\mathbf{51.7}$ & $22.8$          & $14.7$          \\
  & \method                & $71.1$          & $39.9$          & $\mathbf{38.8}$ & $\mathbf{35.0}$ \\
\cmidrule(lr){2-6}
  & \textit{No preference} & --              & --              & $38.4$          & $50.4$          \\
\bottomrule
\end{tabular}
\end{table*}

\FloatBarrier
\clearpage
\section{Controllability of \method}
\label{sec:control}
\label{app:control}

Unlike prompting, where modifiers such as ``more'' or ``less'' yield qualitative, unpredictable shifts~\citep{cheng-genctrl}, steering methods expose an explicit strength parameter $\lambda$. However, $\lambda$ is typically unbounded and therefore difficult to interpret. Following LinEAS~\citep{lineas}, \method adopts an optimal transport formulation that grounds $\lambda$ in a predictable and interpretable $[0, 1]$ range: $\lambda=0$ means no intervention, $\lambda=1$ means full transport from the source to the target distribution, while every intermediate value corresponds to a partial transport that progressively closes the gap between the source and the target distribution.

\begin{figure}[t]
\centering
    \includegraphics[width=0.6\linewidth, trim=0 0 0 0, clip]{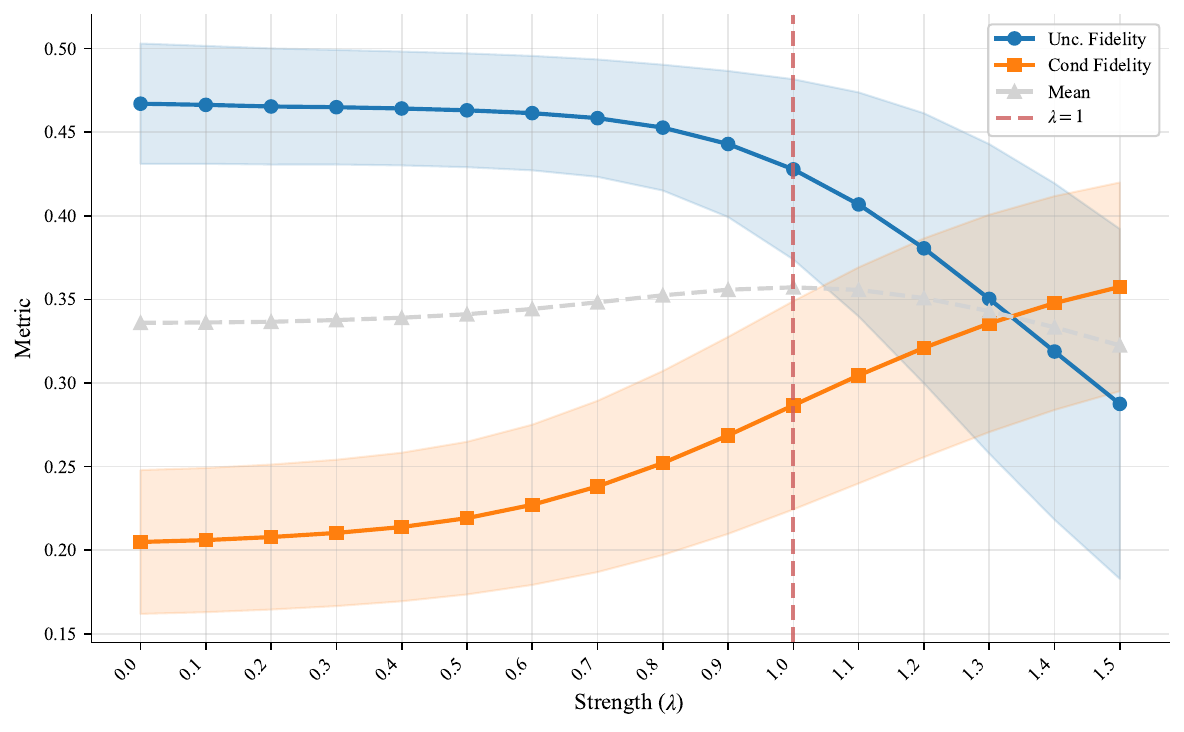}
\caption{\textbf{Controllability of \method.} Increasing the conditioning parameter $\lambda$ increases the conditioning (\conceptfidelityshort) while minimally diverging from the prompt (\promptfidelityshort), achieving the optimal compromise (measured as arithmetic mean) for $\lambda=1$.}
\label{fig:controllability}
\end{figure}

This provides an intuitive ``knob'' for control. As shown in~\cref{fig:controllability}, increasing $\lambda$ monotonically and smoothly increases \conceptfidelity while maintaining \promptfidelity close to its original value. Out of curiosity, we also tested values of $\lambda>1$. As expected \conceptfidelity continues to increase but at the cost of a drop in \promptfidelity. Computing the arithmetic mean between the two metrics shows that indeed the optimal value is reached for $\lambda=1$.

\begin{figure}[h]
    \centering
    \caption{Examples of controlled conditioning using \dmdtwo. The first image for each generation corresponds to $\lambda=0$ (\ie no conditioning), the last to $\lambda=1.5$. The image highlighted in red corresponds to $\lambda=1$.}
    \label{fig:control_qualitative}
    
    \begin{minipage}{\linewidth}
        \raggedright 
        \textbf{Prompt:} \emph{Hands holding a melting ice cream}. \\
        \textbf{Concept:} \emph{Expressionist}
        \vspace{2mm}
        \includegraphics[width=\linewidth]{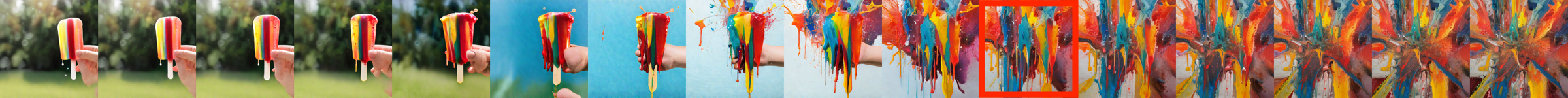}
    \end{minipage}
    
    \vspace{3mm}
    
    \begin{minipage}{\linewidth}
        \raggedright
        \textbf{Prompt:} \emph{Hailstones collecting in a bucket. More falling around.} \\
        \textbf{Concept:} \emph{Contemporary Art}
        \vspace{2mm}
         \includegraphics[width=\linewidth]{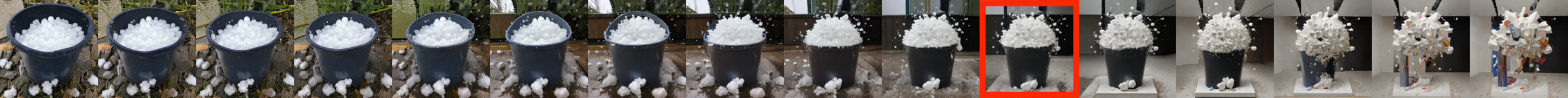}
    \end{minipage}
    
    \vspace{3mm}
    
    \begin{minipage}{\linewidth}
        \raggedright
        \textbf{Prompt:} \emph{A four-leaf clover among thousands of three-leaf ones.} \\
        \textbf{Concept:} \emph{Contemporary Art}
        \vspace{2mm}
         \includegraphics[width=\linewidth]{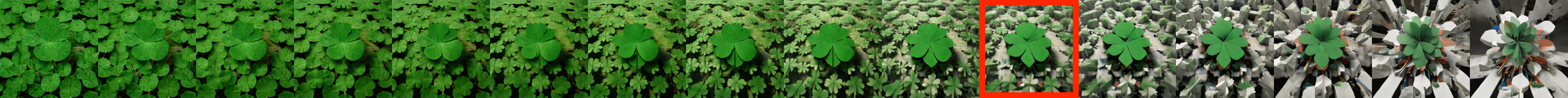}
    \end{minipage}

    \vspace{3mm}

    \begin{minipage}{\linewidth}
        \raggedright
        \textbf{Prompt:} \emph{A tiny mouse peeking out from behind a wheel of cheese.} \\
        \textbf{Concept:} \emph{Art Deco}
        \vspace{2mm}
         \includegraphics[width=\linewidth]{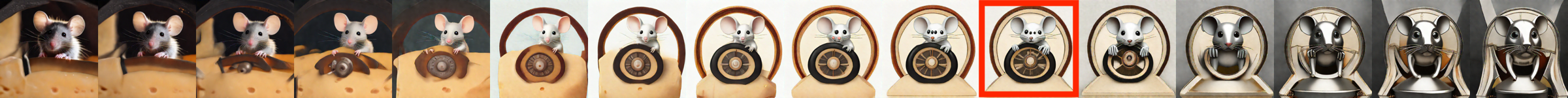}
    \end{minipage}
    
    \vspace{3mm}
    
    \begin{minipage}{\linewidth}
        \raggedright
        \textbf{Prompt:} \emph{Coins stacked in uneven towers of different currencies.} \\
        \textbf{Concept:} \emph{Contemporary Art}
        \vspace{2mm}
         \includegraphics[width=\linewidth]{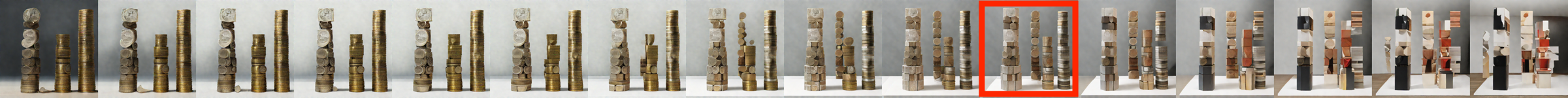}
    \end{minipage}
\end{figure}

\Cref{fig:control_qualitative} shows qualitative examples of this controlled generation on \dmdtwo: we start at $\lambda=0$ (no conditioning) and progressively increase $\lambda$ to $1.5$, with the $\lambda=1$ generation highlighted in red.

\FloatBarrier

\clearpage
\section{Target Concepts}\label{app:targetconcepts}
Below is the full list of 777 concepts used in our studies, organized across 10 categories. The dataset is randomly split into 610 training and 167 test concepts stratifying across categories.

\small

\paragraph{Art Techniques (100).}
abstract-drawing, acrylic, action-painting, airbrush, alcohol-technique, assemblage, automatic-drawing, blending, blind-contour, brush-painting, calligraphy, ceramic-art, charcoal, chiaroscuro, collage, color-mixing, continuous-line, contour-drawing, crackle-medium, cross-hatching, cut-and-paste, dead-coloring, decoupage, divisionism, drip-painting, dry-brush, dry-brush-technique, dry-media, egg-tempera, engraving, etching, fabric-collage, fan-brush, figure-drawing, finger-painting, fluorescent-paint, forced-perspective, fresco, gel-medium, gestural-painting, gesture-drawing, glazing, glowing-paint, gouache, graphite, green-underpainting, impasto, ink-wash, interference-paint, lettering, life-drawing, line-art, linocut, lithography, masking-fluid, matte-painting, metallic-paint, milk-paint, mixed-media-collage, mosaic, neo-impressionism, oil-glazing, oil-painting, optical-illusions, optical-mixing, palette-knife, palette-knife-painting, paper-collage, papier-mache, perspective-illusion, perspective-tricks, photomontage, plastic-wrap-technique, plein-air-painting, pour-painting, relief-painting, resist-technique, salt-technique, sand-texture, scratch-art, screen-print, sketch-art, speed-painting, splattering, sponge-technique, spray-painting, stained-glass, stippling, tempera, texture-paste, thumb-painting, underpainting, unmixed-colors, wash-painting, watercolor, wax-painting, wet-brush, wet-media, wet-on-wet, woodcut.

\paragraph{Artistic Movements (47).}
abstract-expressionist, art-deco, art-nouveau, arte-povera, avant-garde, baroque, bauhaus, conceptual-art, constructivist, contemporary-art, cubist, dadaist, de-stijl, expressionist, fauvism, fluxus, futurist, geometric-abstract, gothic-revival, impressionist, installation-art, kinetic-art, land-art, magic-realism, mannerism, minimalist, modernist, muralism, neo-baroque, neo-expressionist, neoclassical, neue-sachlichkeit, op-art, photorealistic, pointillism, pop-art, post-impressionist, psychedelic, realism, renaissance, rococo, romanticism, street-art, suprematist, surrealist, transavantgarde, ukiyo-e.

\paragraph{Color Treatments (80).}
analogous-colors, atmospheric-glow, bleach-bypass, bleached, blue-highlights, blue-hour-light, bold-contrast, brass-finish, bronze-tint, cinematic-grade, cinematic-lighting, color-grading, color-pop, color-wash, complementary-colors, cool-palette, cool-shadows, copper-patina, copper-tint, cyberpunk-lighting, deep-shadows, desaturated-tones, digital-display, dramatic-lighting, duotone, earth-palette, earth-tones, faded, film-look, fluorescent, gentle-highlights, god-rays, golden-hour-light, golden-tint, gradient-map, holographic-color, iridescent, jewel-tones, metallic-tones, midnight-colors, monochromatic-colors, monochrome, muted, muted-pastels, neon, ocean-palette, orange-teal, pastel, pearlescent, psychedelic-color, rainbow-gradient, retro-computing, retro-filter, retro-tint, rich-contrast, rim-lighting, rust-tint, saturated-colors, screen-glow, screen-reflection, selective-color, sepia, sharp-contrast, silver-shine, silver-tint, soft-lighting, split-toning, subtle-contrast, sunrise-colors, sunset-colors, synthwave-colors, twilight-colors, vaporwave-aesthetic, vibrant, vintage-color, vivid-highlights, vivid-saturation, volumetric-lighting, warm-palette, warm-shadows.

\paragraph{Digital Methods (44).}
algorithmic-art, ambient-light, ascii-art, atmospheric-haze, augmented-reality, backlighting, bitmap, blender-render, cel-shading, cgi, chrome-reflection, color-fringing, color-temperature, cyber-aesthetic, data-visualization, digital-illustration, digital-painting, flat-shading, fractal, generative-art, glass-distortion, glass-material, glitch-art, high-poly, hologram-effect, holographic-art, isometric-view, low-poly, motion-graphics, net-art, octane-render, particle-effects, photo-manipulation, photorealistic-render, pixel-art, procedural-art, ray-tracing, toon-shader, unreal-engine, vector-art, vector-graphics, virtual-reality, voxel-art, wireframe-render.

\paragraph{Environmental Settings (124).}
abandoned-city, abandoned-factory, airport, alien-planet, alleyway, ancient-ruins, ancient-temple, arctic, arctic-station, art-gallery, autumn-forest, ballroom, bamboo-grove, beach-house, bridge, calm-lake, canyon, castle-ruins, cathedral, cave, cherry-blossom, coastal, concert-hall, conservatory, coral-reef, countryside, crystal-cave, crystal-palace, deep-forest, deep-sea, desert, desert-camp, desert-landscape, digital-dystopia, downtown, enchanted-garden, farmland, floating-city, floating-islands, flower-field, foggy-moor, forest, forest-clearing, fortress, frozen-lake, futuristic-city, ghost-town, glacier, grand-hotel, greenhouse, highway, hot-spring, industrial-zone, jungle-camp, kelp-forest, laboratory, lava-tube, library, lighthouse, lighthouse-cliff, lush-jungle, magical-forest, medieval-castle, misty-valley, monastery, moon-base, mountain-lodge, mountain-vista, mountainous, mystical-grove, neon-alley, neon-cityscape, neon-metropolis, neon-tokyo, night-market, oasis, observatory, ocean-view, old-town, orchard, railway, rainforest, rainy-street, research-station, retro-diner, rooftop, rushing-river, sand-dunes, savanna, seaport, shelter, shopping-arcade, sky-city, snowy-peaks, space, space-station, speculative-future, spring-meadow, steampunk-factory, stormy-sea, stratosphere, street-festival, subterranean, suburb, summer-beach, summer-cottage, suspension-bridge, theater-stage, train-station, tropical, tundra, tunnel, underwater, underwater-city, urban, vineyard, vintage-diner, volcanic, volcano-observatory, watchtower, wilderness, windy-cliff, winter-cabin, winter-wonderland.

\paragraph{Fantasy Genres (77).}
80s-aesthetic, academic-aesthetic, adventure-game, alternate-history, arthurian, atompunk, aurora-aesthetic, biopunk, brutalist-aesthetic, cosmic-aesthetic, cottagecore, crystal-core, cybernetic-style, cyberpunk, cyberpunk-aesthetic, dark-electronic, dark-fantasy, decora-kei, dieselpunk, dreamscape, dreamy-electronic, dystopian, epic-fantasy, fairy-core, fairy-tale-retelling, fantasy-realm, folkloric, forest-core, future-funk, galactic-aesthetic, gem-core, gothic-fantasy, gothic-lolita, harajuku, hard-sci-fi, heroic-fantasy, high-fantasy, holographic-aesthetic, iridescent-aesthetic, liminal-space, magical-realism, marble-aesthetic, medieval-fantasy, military-aesthetic, modern-fantasy, multiverse, mystical, nature-spirituality, neo-noir, neon-noir, ocean-core, otherworldly, parallel-universe, paranormal-romance, prismatic-aesthetic, psychedelic, psychedelic-tech, retro-futurism, romantic-gothic, science-fantasy, sky-core, soft-sci-fi, solarpunk, space-opera, speculative-fiction, spiritual, steampunk, superhero-style, surreal-fantasy, sword-and-sorcery, synthwave, tech-noir, time-travel, toy-render, urban-fantasy, utopian, vaporwave.

\paragraph{Historical Periods (99).}
1920s, 1950s, 1960s, 1970s, 1980s, 1990s, ai-era, ancient-egypt, ancient-greece, ancient-rome, art-deco-era, art-nouveau-period, atomic-age, atomic-future, aztec, baroque-era, belle-epoque, biotech-era, bronze-age, byzantine-empire, chrome-age, chrome-era, classical-antiquity, climate-change-era, cold-war-era, colonial, computer-age, computer-revolution, counter-reformation, cyberpunk-era, dark-ages, digital-age, dot-com-era, early-medieval, early-renaissance, edwardian, electric-age, enlightenment, folk-tradition, galactic-era, gilded-age, gothic-period, great-depression, grunge-nineties, hellenistic, heritage-style, high-medieval, high-renaissance, incan, industrial-era, internet-age, interplanetary-age, iron-age, jazz-age, jet-age, late-medieval, machine-age, mannerism, mars-colonization, mayan, mid-century-modern, millennium-era, modernist-period, nanotech-era, neoclassical-period, neolithic, neon-eighties, neon-era, northern-renaissance, nuclear-age, paleolithic, pandemic-era, post-war-1950s, prehistoric, psychedelic-seventies, quantum-age, renaissance-era, renewable-energy-era, retro-futuristic, retro-gaming, rococo-period, roman-empire, romanesque, romantic-period, smartphone-era, social-media-era, social-network-era, space-age, space-exploration, steam-age, streaming-era, swinging-sixties, television-age, victorian, viking, vintage-computing, vintage-revival, wartime-1940s, y2k-era.

\paragraph{Illustration Styles (64).}
3d-illustration, abstract-illustration, acrylic-illustration, album-cover, anime-style, architectural-drawing, art-deco-poster, book-cover, botanical-illustration, brushwork, caricature, cartoon, character-design, charcoal-illustration, children's-book, colored-sketch, comic-book, concept-art, costume-design, creature-design, decorative, digital-sketch, disney-style, editorial-illustration, environment-design, fairy-tale, fantasy-illustration, fine-art, flat-colors, geometric-illustration, gouache-illustration, graphic-novel, illustrated-book, ink-drawing, isometric-illustration, japanese-animation, kawaii-style, korean-pop-style, magazine-illustration, manga, map-illustration, marker-sketch, maximalist, miyazaki-style, modern-illustration, newspaper-cartoon, oil-illustration, pastel-illustration, pen-sketch, pixar-style, prop-design, refined-sketch, retro-illustration, rough-sketch, sci-fi-illustration, sketch-style, studio-ghibli-style, traditional-illustration, travel-poster, vector-illustration, vehicle-design, vintage-poster, watercolor-illustration, web-illustration.

\paragraph{Individual Artists (61).}
alex-katz, alphonse-mucha, amy-sherald, andrew-wyeth, annie-leibovitz, anselm-kiefer, artgerm, banksy, basquiat, bourgeois, caravaggio, cindy-sherman, cy-twombly, da-vinci, dali, david-hockney, diane-arbus, edward-hopper, francis-bacon, georgia-okeefe, gerhard-richter, greg-rutkowski, haring, helen-frankenthaler, helmut-newton, hokusai, irving-penn, jasper-johns, kahlo, kara-walker, katsuhiro-otomo, kehinde-wiley, klimt, kusama, lucian-freud, magritte, makoto-shinkai, mamoru-hosoda, michelangelo, monet, munch, norman-rockwell, peter-halley, picasso, pollock, rembrandt, richard-avedon, richard-diebenkorn, robert-rauschenberg, rothko, satoshi-kon, takashi-murakami, takeshi-obata, tim-burton, van-gogh, vermeer, vivian-maier, warhol, wayne-thiebaud, wes-anderson, willem-de-kooning.

\paragraph{Photography \& Cinema (80).}
16mm-film, 35mm-film, 360-photography, 3d-photography, 4k-cinema, 70mm-film, 8k-cinema, 8mm-film, aerial-photography, analog-photography, anamorphic, architectural-photography, astrophotography, black-and-white, blurred-background, bokeh-background, boudoir-photography, camera-obscura, cinematic, cross-processed, cross-processing, digital-cinema, digital-photography, documentary, double-exposure, drone-photography, event-photography, fashion-photography, film-grain, film-noir, film-photography, fisheye-lens, food-photography, glamour-photography, headshot-photography, high-speed-photography, imax, infrared-photography, instant-film, instant-photography, interior-photography, kodachrome, landscape, large-format, light-painting, lomographic, lomography, long-exposure, macro-photography, medium-format, microscopy, milky-way-photography, multiple-exposure, nature-photography, pinhole-camera, polaroid-style, portrait, product-photography, sepia-tone, slide-film, star-trails, steel-wool-photography, stereoscopic, still-life-photography, stop-motion, street-photography, studio-lighting, super-8-film, technicolor, telephoto-lens, tilt-shift, time-lapse, ultraviolet-photography, underwater-photography, vintage-photography, vr-photography, wedding-photography, wide-angle-lens, wildlife-photography, x-ray-photography.

\normalsize

\FloatBarrier

\newcommand{\Himg}{2.6cm}
\newcommand{\Worig}{0.23\textwidth}
\newcommand{\Wsign}{0.03\textwidth}
\newcommand{\Wcond}{0.08\textwidth}
\newcommand{\Wimgc}{0.175\textwidth}

\newcommand{\DEBUGCELLS}{0}
\newcommand{\maybeFbox}[1]{\ifnum\DEBUGCELLS=1 \fbox{#1}\else #1\fi}

\newcommand{\imgcell}[1]{%
  \maybeFbox{%
    \parbox[c][\Himg][c]{\linewidth}{\centering
      \includegraphics[height=\Himg,width=\linewidth,keepaspectratio]{#1}}%
  }%
}
\newcommand{\textcell}[1]{%
  \maybeFbox{\parbox[c][\Himg][c]{\linewidth}{\centering #1}}%
}
\newcommand{\pluscell}{\textcell{\Large\bfseries +}}
\newcommand{\eqcell}{\textcell{\Large\bfseries =}}

\newcommand{\QualRow}[5]{%
  \imgcell{#1} & \pluscell & \textcell{\scriptsize\bfseries #2} & \eqcell &
  \imgcell{#3} & \imgcell{#4} & \imgcell{#5} \\[6pt]
}

\FloatBarrier
\clearpage
\section{Qualitative examples}
\label{app:qualitative}

We show here qualitative examples of different conditioning methods on DMD2.

\begin{figure*}[h]
\centering
\setlength{\tabcolsep}{10pt}
\renewcommand{\arraystretch}{1.2}
\begin{tabular}{m{0.12\textwidth} c m{0.12\textwidth} c m{0.12\textwidth} m{0.12\textwidth} m{0.12\textwidth}}
\centering\textbf{Original} & & \centering\textbf{Conditioning} & & \centering\textbf{Prompting} & \centering\textbf{LinEAS} & \centering\textbf{HyperTransport}\tabularnewline[5pt]

\centering\includegraphics[width=0.12\textwidth]{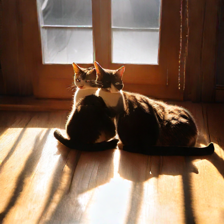} & 
\Large + & 
\centering deep-sea & 
\Large = & 
\centering\includegraphics[width=0.12\textwidth]{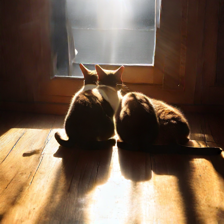} &
\centering\includegraphics[width=0.12\textwidth]{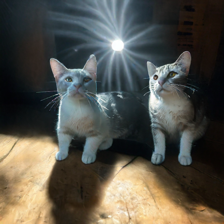} &
\centering\includegraphics[width=0.12\textwidth]{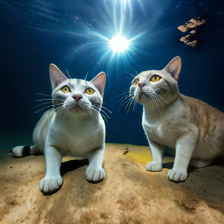} \tabularnewline[8pt]

\centering\includegraphics[width=0.12\textwidth]{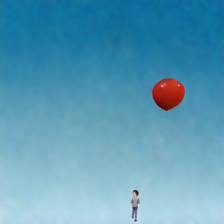} & 
\Large + & 
\centering Monet & 
\Large = & 
\centering\includegraphics[width=0.12\textwidth]{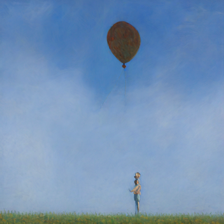} &
\centering\includegraphics[width=0.12\textwidth]{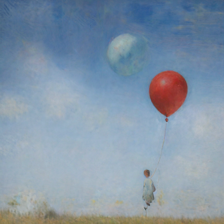} &
\centering\includegraphics[width=0.12\textwidth]{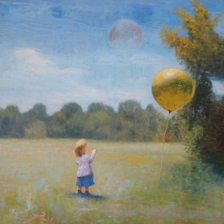} \tabularnewline[8pt]

\centering\includegraphics[width=0.12\textwidth]{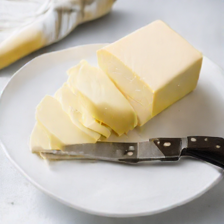} & 
\Large + & 
\centering digital-illustration & 
\Large = & 
\centering\includegraphics[width=0.12\textwidth]{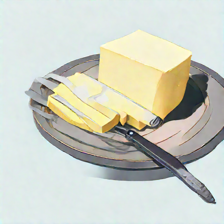} &
\centering\includegraphics[width=0.12\textwidth]{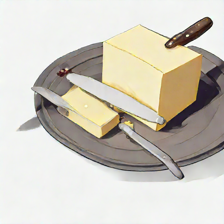} &
\centering\includegraphics[width=0.12\textwidth]{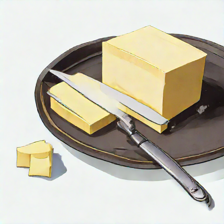} \tabularnewline[8pt]

\centering\includegraphics[width=0.12\textwidth]{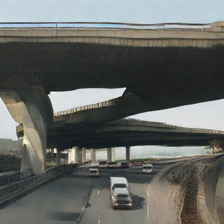} & 
\Large + & 
\centering {\shortstack[c]{green-\\underpainting}} & 
\Large = & 
\centering\includegraphics[width=0.12\textwidth]{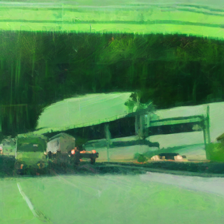} &
\centering\includegraphics[width=0.12\textwidth]{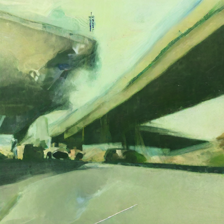} &
\centering\includegraphics[width=0.12\textwidth]{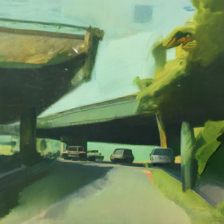} \tabularnewline[8pt]

\centering\includegraphics[width=0.12\textwidth]{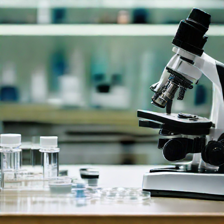} & 
\Large + & 
\centering {\shortstack[c]{cosmic-\\aesthetic}} & 
\Large = & 
\centering\includegraphics[width=0.12\textwidth]{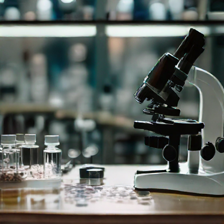} &
\centering\includegraphics[width=0.12\textwidth]{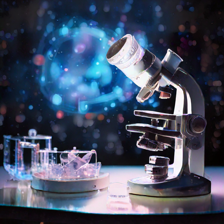} &
\centering\includegraphics[width=0.12\textwidth]{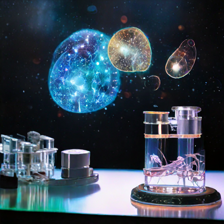} \tabularnewline[8pt]

\end{tabular}
\caption{Qualitative comparison of conditioning methods on \dmdtwo. Original image (generated with the original prompt), conditioning concept, and outputs from three methods. The last row shows a failure case of \method, where the microscope has been removed in favor of conditioning more on the style.}
\label{fig:qualitative_comparison}
\end{figure*}

\FloatBarrier
\clearpage

\section{VLM-as-a-Judge Evaluation}\label{app:vlm_study}

We evaluate \method in a VLM-as-a-judge setting using \vlmmodel~\citep{chen2024internvl} on $33{,}400$ test image pairs (text-conditioned and image-conditioned, summed across both models) covering all test concepts. A two-stage protocol decouples input fidelity from concept fidelity.

\paragraph{Stage 1: Input fidelity.} Each image is judged independently for faithfulness to the original prompt, regardless of applied style. The VLM prompt reads:

\begin{quote}
\small\ttfamily
``Look at this image. It was generated from this prompt: ``\{PROMPT\}'' Reason first. Is this image faithful to the prompt? Note: The image has been stylized using the ``\{CONCEPT\}'' style. When judging faithfulness, focus on whether the semantic content (objects, actions, relationships, counts, etc.) matches the prompt, regardless of the artistic style applied. Respond with ONLY: YES or NO''
\end{quote}

\paragraph{Stage 2: Concept fidelity.} Images are shown side-by-side and the judge selects which better represents the target concept:

\begin{quote}
\small\ttfamily
``Look at this image showing two pictures arranged \{pos\_first\} and \{pos\_second\}. Both pictures should display this concept/style: ``\{CONCEPT\}'' Reason first. Which image better displays the concept? Respond with ONLY: \{pos\_first\}, \{pos\_second\}, BOTH (if equally good), or NONE (if neither displays the concept well)''
\end{quote}

To control for position bias, each comparison runs twice with swapped positions; only position-agnostic answers are retained. We report concept preference among prompt-faithful pairs.

\paragraph{Results.} Prompting more often produces prompt-faithful images ($82.9\%$ vs.\ ${\sim}71\%$), but with weaker concept induction. Among the faithful subset ($15{,}204$ text-conditioned pairs; $15{,}426$ image-conditioned pairs), \method is preferred $42.9\%$ vs.\ $21.6\%$ for text-conditioning, and $39.6\%$ vs.\ $23.5\%$ for image-conditioning.

\paragraph{Relation to the user study.} The VLM-as-a-judge protocol above mirrors the two-stage design of the user study (\Cref{app:userstudy}), which serves as the reference protocol: Stage~1 corresponds to the single-image \emph{Prompt Faithfulness} stage (``Does the image generated from the prompt faithfully depict the original prompt?''), and Stage~2 to the pairwise \emph{Concept Comparison} stage (``Which image better displays the concept?''). The VLM prompts above are the per-instance instructions presented to the judge model; the instructions presented to human judges are reproduced verbatim in \Cref{app:userstudy}.

\begin{figure}[H]
    \centering
    \includegraphics[width=\linewidth]{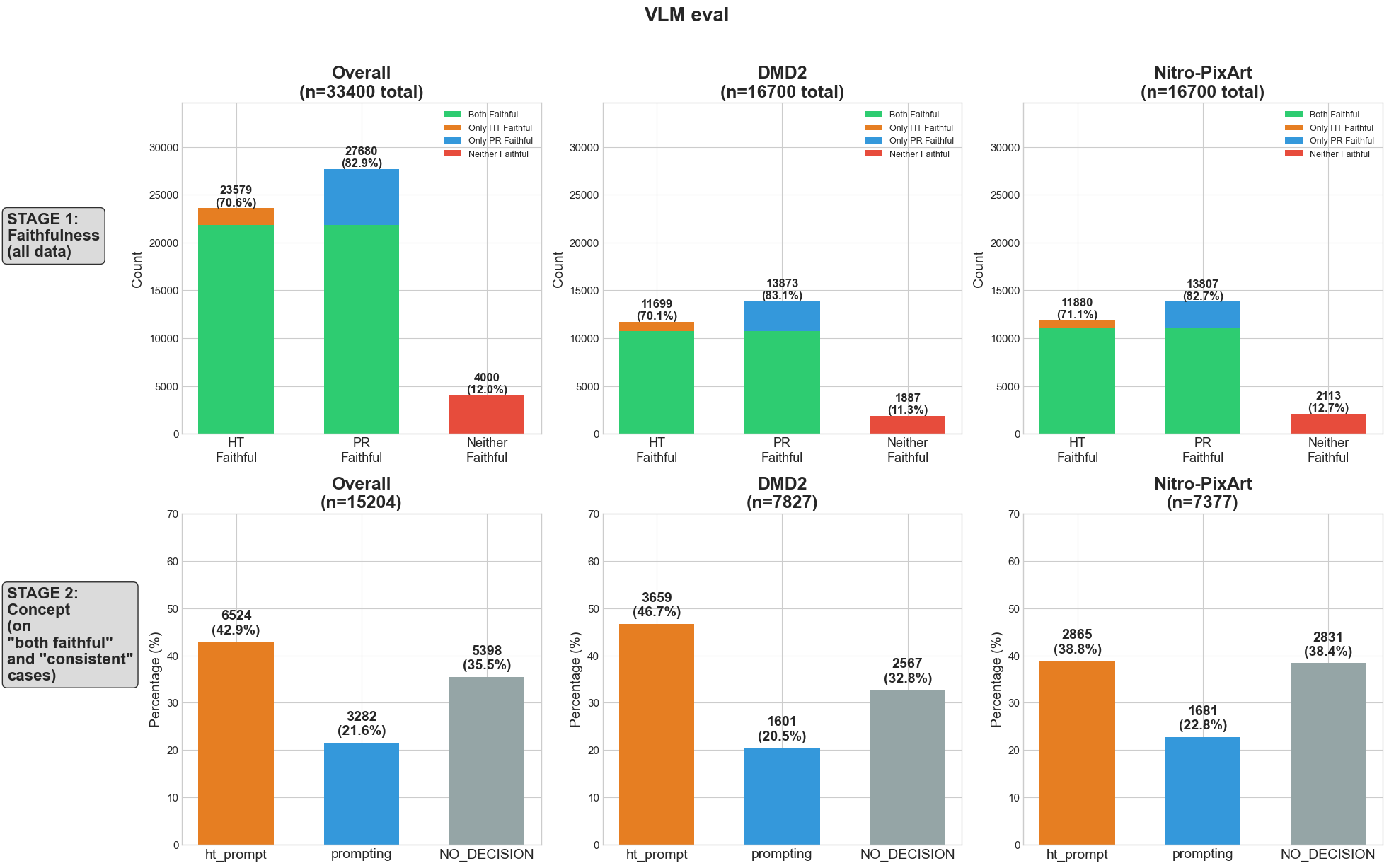}
    \caption{\textbf{Two-stage evaluation of images generated via text-conditioned \method (Test).} As expected, input-fidelity trades-off concept-fidelity. \textbf{Top: Stage 1, Input Fidelity (images judged one by one).} Distribution of faithfulness judgments across all samples and per model (\dmdtwo, \nitro). Prompting produces images that remain more faithful to the original input (${\sim}83\%$ \textit{vs.}\ ${\sim}70\%$ for \method). \textbf{Bottom: Stage 2, Concept fidelity (images judged in pairs).} Among samples where both models produce faithful outputs, \method generations are favored ($42.9\%$ \textit{vs.}\ $21.6\%$). Results are consistent across the two models, with \dmdtwo yielding slightly stronger preferences for \method ($46.7\%$ \textit{vs.}\ $38.8\%$).}
    \label{fig:vlm_text_cond}
\end{figure}

\begin{figure}[H]
    \centering
    \includegraphics[width=\linewidth]{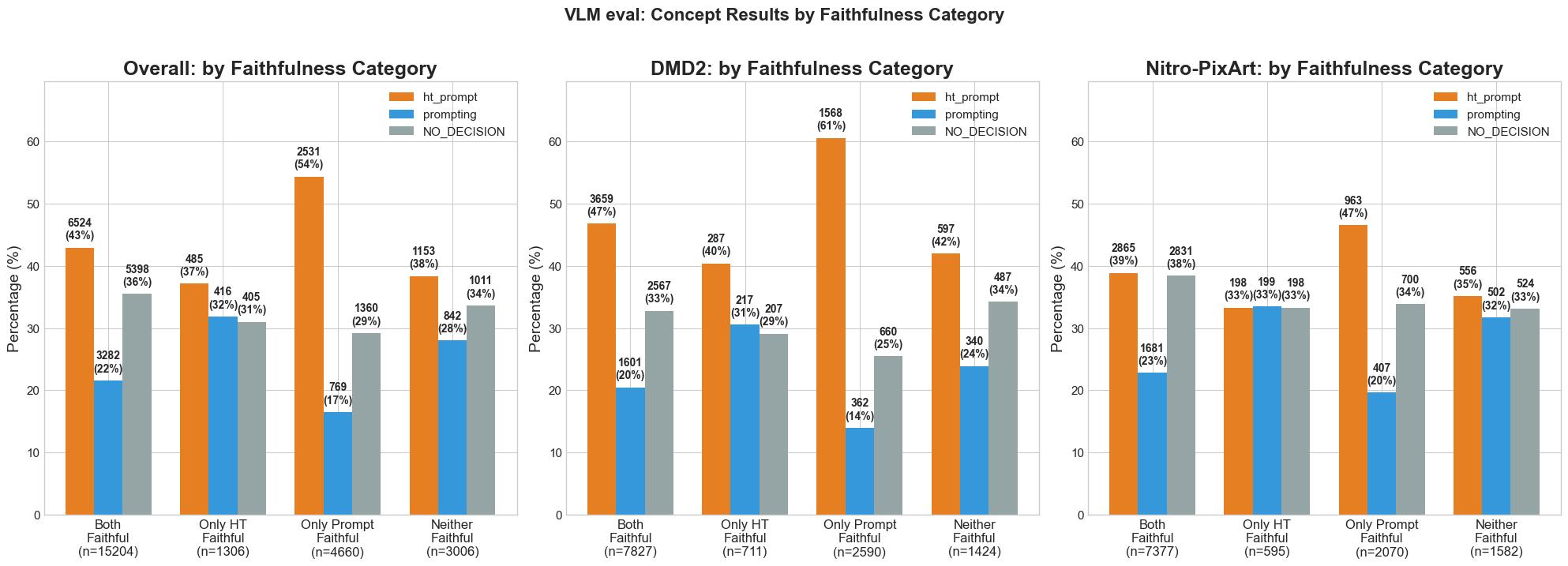}
    \caption{\textbf{Stage 2 (concept-fidelity) broken down by results of Stage 1 (input-fidelity)} (HT text-conditioned, Test). \method generations are preferred irrespectively of their ``faithfulness'' category, confirming that the concept-fidelity advantage of \method is not merely a byproduct of input-faithful samples. Results hold across models.}
    \label{fig:vlm_text_cond_by_category}
\end{figure}

\FloatBarrier
\clearpage

\section{Concept Distribution Analysis}\label{app:concept_distribution}

To verify that the random concept split is balanced and that test concepts are not simply interpolations of training concepts, we compute mean CLIP embeddings per concept over each sentence and analyse the resulting cosine-distance structure between splits.

\begin{figure}[H]
    \centering
    \includegraphics[width=\linewidth]{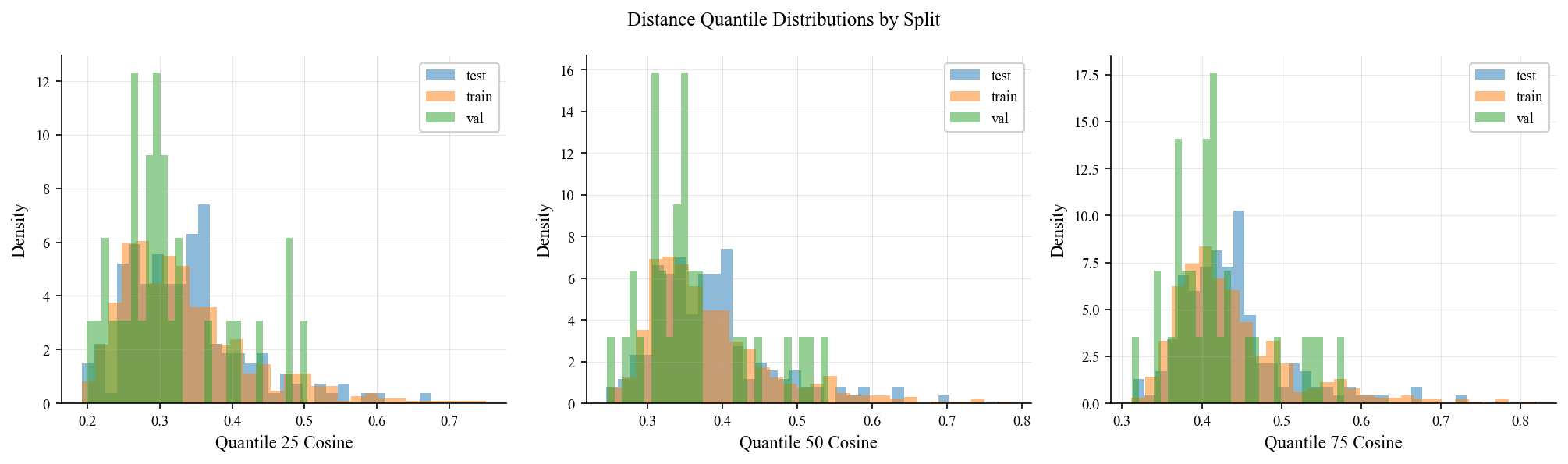}
    \caption{\textbf{The random split is balanced across train/val/test.} For each concept we compute pairwise CLIP cosine distances to the rest of the dataset and summarise them by the $25$th, $50$th and $75$th percentile (left, middle, right panel). The three splits occupy nearly identical regions of each quantile axis, indicating that the random split does not concentrate the easier or harder concepts in any one split.}
    \label{fig:concept_split_balance}
\end{figure}

\paragraph{Split balance.} For every concept we summarise its pairwise CLIP cosine distances to the rest of the dataset by the $25$th, $50$th and $75$th percentile, and plot the distribution of those quantiles across concepts in each split (\Cref{fig:concept_split_balance}). Train, val and test occupy near-identical regions of each quantile axis, indicating that the random split does not concentrate the easier or harder concepts in any one split.

\begin{figure}[H]
    \centering
    \includegraphics[width=0.48\textwidth]{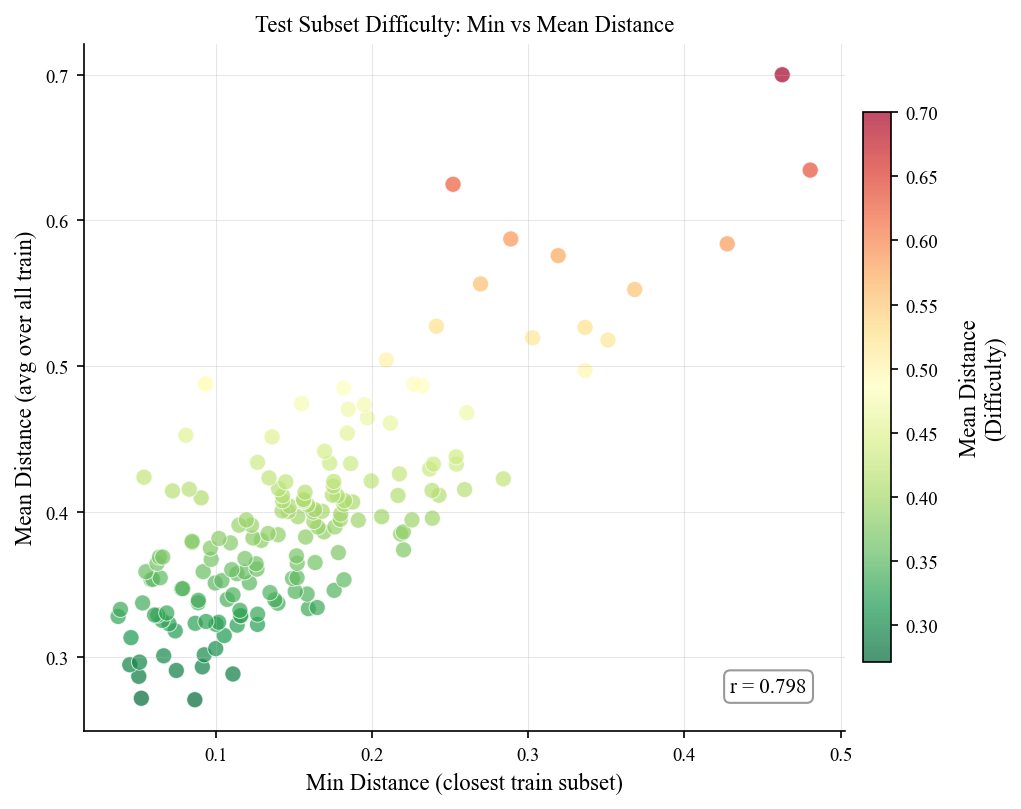}\hfill
    \includegraphics[width=0.48\textwidth]{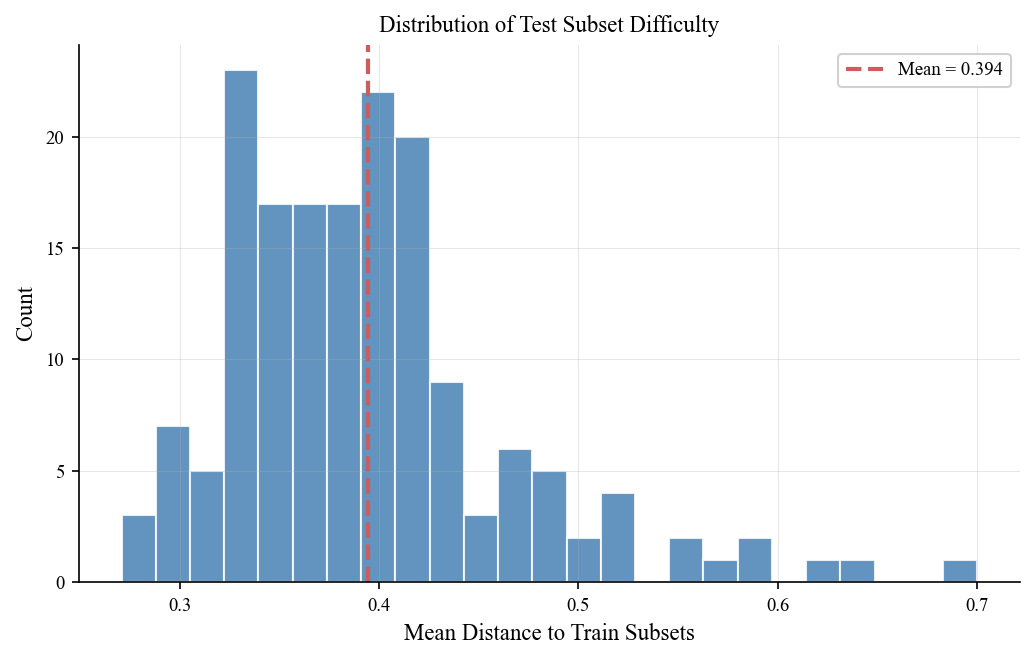}
    \caption{\textbf{Test concepts span a wide range of distances from training.} \textbf{Left:} per-test-concept mean cosine distance to all training concepts (y) vs.\ minimum cosine distance to the closest training concept (x); the strong correlation ($r{=}0.80$) confirms that the mean-distance proxy reflects both nearest-neighbor and global novelty. \textbf{Right:} distribution of test-concept difficulty (mean cosine distance to training concepts); difficulties span ${\sim}0.27$ to ${\sim}0.70$ (mean $0.394$), with substantial mass at distances well above $0.4$. The consistent performance of \method across this range (\Cref{tab:main_comparison}) confirms that test results are not driven by interpolation.}
    \label{fig:concept_difficulty}
\end{figure}

\paragraph{Difficulty proxy.} We then quantify the difficulty of a test concept as the mean CLIP cosine distance of its embedding to all training-concept embeddings. \Cref{fig:concept_difficulty} (left) validates this proxy: it correlates strongly (Pearson $r{=}0.80$) with the minimum distance to the closest training neighbor, so it captures both global novelty and nearest-neighbor proximity rather than averaging them away.

\paragraph{Difficulty distribution.} \Cref{fig:concept_difficulty} (right) shows the distribution of this difficulty across test concepts: it spans ${\sim}0.27$ to ${\sim}0.70$ (mean $0.394$), with substantial mass beyond $0.4$, i.e., many test concepts lie far from every training neighbor. The consistent performance of \method across this full range (\Cref{tab:main_comparison}) therefore reflects genuine generalization rather than interpolation.

\FloatBarrier
\clearpage

\section{Training Curves: MLP vs.\ PerceiverIO}\label{app:training_curves}

Both the MLP and PerceiverIO architectures converge with comparable loss profiles when trained with our 1D Wasserstein alignment loss, confirming the stability of the objective across architectures. The MLP achieves marginally better downstream metrics and was selected for simplicity.

\begin{figure}[H]
    \centering
    \includegraphics[width=0.48\textwidth]{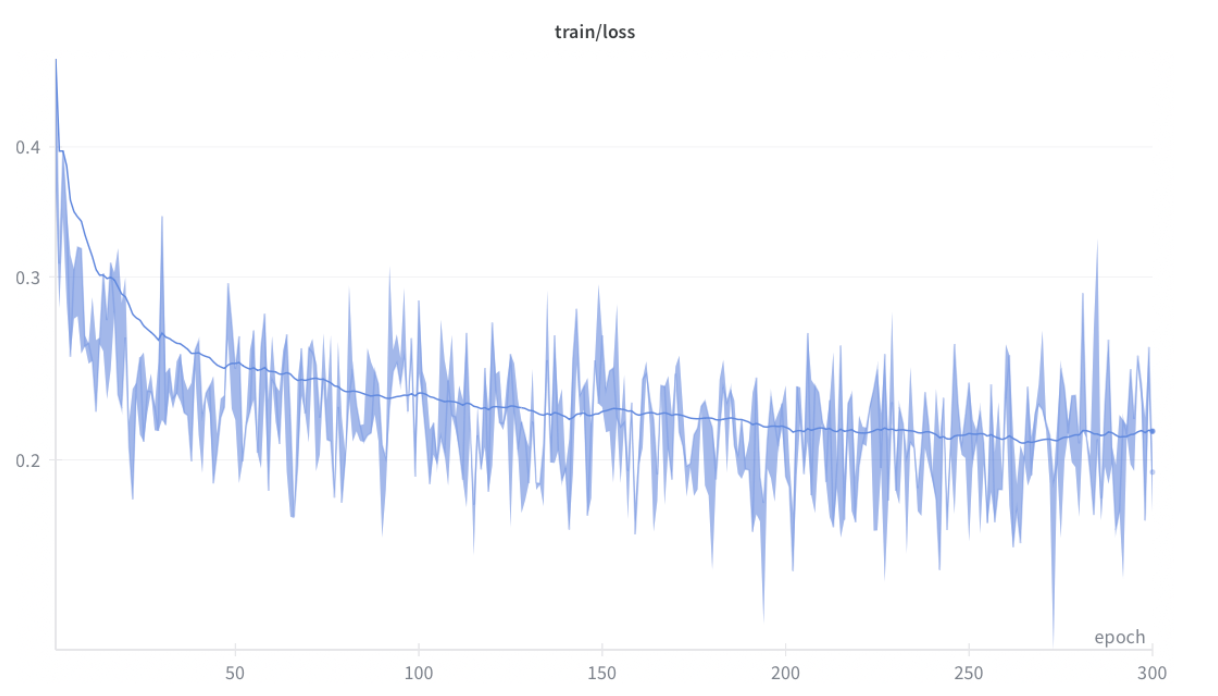}\hfill
    \includegraphics[width=0.48\textwidth]{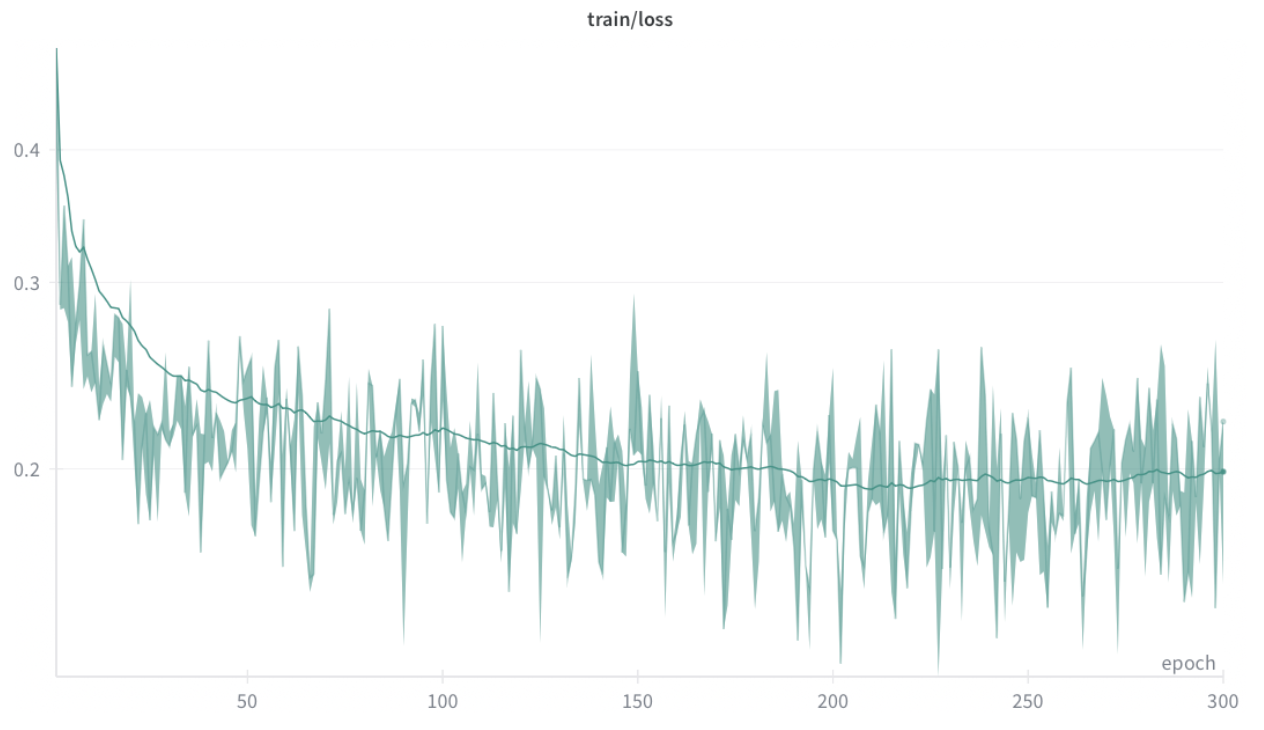}
    \caption{\textbf{Training curves of Perceiver and MLP architectures.} Training loss over 300 epochs for the Perceiver (left) and MLP (right). Both architectures converge to similar final loss values (${\sim}0.2$), with the MLP exhibiting slightly lower variance in later epochs.}
    \label{fig:training_curves}
\end{figure}

\applefootnote{\textcolor{textgray}{\sffamily Apple and the Apple logo are trademarks of Apple Inc., registered in the U.S. and other countries and regions.}}

\end{document}